%% file: acl_latex.tex
\pdfoutput=1

\documentclass[11pt]{article}

\usepackage[final]{acl}

\usepackage{times}
\usepackage{latexsym}
\usepackage{amsmath}
\usepackage{adjustbox}
\usepackage{graphicx}
\usepackage{booktabs}
\usepackage{multirow}
\usepackage{multicol}
\usepackage{makecell}
\usepackage{enumitem}
\usepackage{arydshln}
\usepackage{xspace}
\usepackage{amssymb}
\usepackage{pifont}
\usepackage{hyperref}
\usepackage[symbol]{footmisc}

\usepackage[T1]{fontenc}

\usepackage[utf8]{inputenc}

\usepackage{microtype}

\usepackage{inconsolata}

\usepackage{graphicx}

\newcommand{\musique}{MuSiQue}
\newcommand{\twowiki}{2WikiMultiHopQA}
\newcommand{\hotpot}{HotpotQA}
\newcommand{\llama}{Llama-3.1-8B-Instruct}
\newcommand{\llamal}{Llama-3.1-70B-Instruct}
\newcommand{\qwen}{Qwen2.5-7B-Instruct}
\newcommand{\gpt}{GPT-4o mini}

\title{\textsc{LongFaith}: Enhancing Long-Context Reasoning in LLMs with Faithful Synthetic Data}

\author{
 \textbf{ Cehao Yang\textsuperscript{1,2}}\footnotemark[1],
 \textbf{ Xueyuan Lin\textsuperscript{1,2,4}}\footnotemark[1],
 \textbf{ Chengjin Xu\textsuperscript{1,3}}\footnotemark[1],
 \textbf{ Xuhui Jiang\textsuperscript{1,3}},\\
 \textbf{ Shengjie Ma\textsuperscript{1}},
 \textbf{ Aofan Liu\textsuperscript{1}},
 \textbf{ Hui Xiong\textsuperscript{2}}\footnotemark[2],
 \textbf{ Jian Guo\textsuperscript{1}}\footnotemark[2]
\\
 \textsuperscript{1}IDEA Research, International Digital Economy Academy
 \\
 \textsuperscript{2}Hong Kong University of Science and Technology (Guangzhou)
\\
 \textsuperscript{3}DataArc Tech Ltd.
 \\
 \textsuperscript{4}Hithink RoyalFlush Information Network Co., Ltd
 \\
\texttt{\{yangcehao,linxueyuan,xuchengjin,jiangxuhui,mashengjie,liuaofan,guojian\}@idea.edu.cn}
\\
\texttt{\{cyang289,xlin058\}@connect.hkust-gz.edu.cn}, \texttt{xionghui@ust.hk}\\
}

\begin{document}
\maketitle
\footnotetext[1]{Equal contribution.}
\footnotetext[2]{Corresponding authors.}
\footnotetext[3]{Our code implementation and datasets can be accessed at \href{https://github.com/IDEA-FinAI/LongFaith}{https://github.com/IDEA-FinAI/LongFaith}.}
\input{abs}
\input{intro}
\input{related}
\input{method}

\input{exp}
\input{conclusion}

\section*{Limitations}
While \textsc{LongFaith} demonstrates significant improvements in long-context reasoning tasks, its scalability and generalization to other LLMs remain an open question. Our experiments focused on a single model, and thus, the performance of \textsc{LongFaith} on other general-purpose LLMs still needs further validation. Additionally, while the synthesized instruction sets with lengths of approximately 10,000 tokens successfully generalized to long-context reasoning tasks, future work will explore the extension of \textsc{LongFaith} to generate instructions with even longer contexts and evaluate the impact on model performance. Finally, \textsc{LongFaith} currently concentrates on reasoning tasks, and we plan to explore its generalization to other tasks such as summarization, dialogue generation, and others, to assess its broader applicability.

\bibliography{custom}

\appendix
\onecolumn
\input{app}

\end{document}

%% file: abs.tex
\begin{abstract}

Despite the growing development of long-context large language models (LLMs), data-centric approaches relying on synthetic data have been hindered by issues related to faithfulness, which limit their effectiveness in enhancing model performance on tasks such as long-context reasoning and question answering (QA). These challenges are often exacerbated by misinformation caused by lack of verification, reasoning without attribution, and potential knowledge conflicts. We propose \textsc{LongFaith}, a novel pipeline for synthesizing faithful long-context reasoning instruction datasets. By integrating ground truth and citation-based reasoning prompts, we eliminate distractions and improve the accuracy of reasoning chains, thus mitigating the need for costly verification processes. We open-source two synthesized datasets—\textsc{LongFaith}-SFT and \textsc{LongFaith}-PO—which systematically address multiple dimensions of faithfulness, including verified reasoning, attribution, and contextual grounding. Extensive experiments on multi-hop reasoning datasets and LongBench demonstrate that models fine-tuned on these datasets significantly improve performance. Our ablation studies highlight the scalability and adaptability of the \textsc{LongFaith} pipeline, showcasing its broad applicability in developing long-context LLMs.

\end{abstract}

%% file: intro.tex
\section{Introduction}
\label{sec:intro}

Long-context processing ability has emerged as a significant challenge for large language models (LLMs)~\cite{distracted, lostinthemiddle, irrelevant, sametaskmoretokens}, especially arises when models process extensive textual information, making it hard to recognize relevant evidence and address downstream tasks such as question answering (QA), summarization, and complex reasoning~\cite{longbench, longbenchv2, bench, ruler, helmet}. A variety of model-centric methods have been proposed to extend the length of context windows in LLMs~\cite{extending, longlora, yarn, lminfinite, longrope}. Additionally, many data-centric methods, such as synthesizing long-context understanding instructions for fine-tuning, have gained attention for enhancing models’ ability to handle and utilize extended contexts~\cite{effective, in2, dataengineering, longalign, coc, prolong, longmit, longreward, sealong, sog}.

\begin{figure}[t]
    \centering
    \includegraphics[width=1\linewidth]{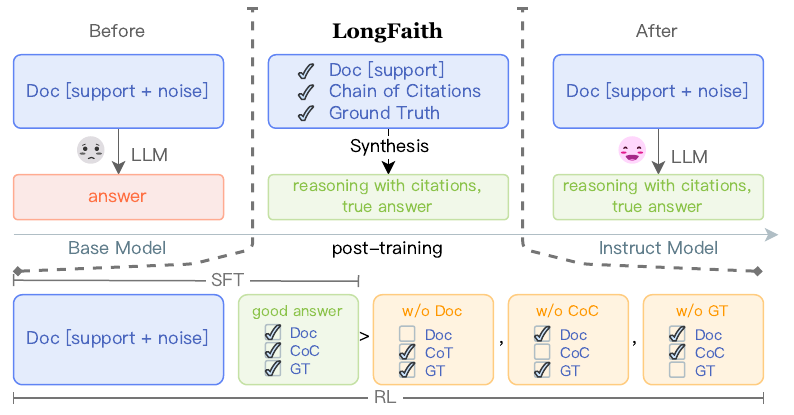}
    \caption{A brief introduction of \textsc{LongFaith}. Synthesized long-context reasoning instruction sets and preference datasets are fed into the post-training stage of downstream LLMs.}
    \label{fig:small}
    \vspace{-5pt}
\end{figure}

\begin{figure*}[t]
    \centering
    \centerline{\includegraphics[width=0.88\linewidth]{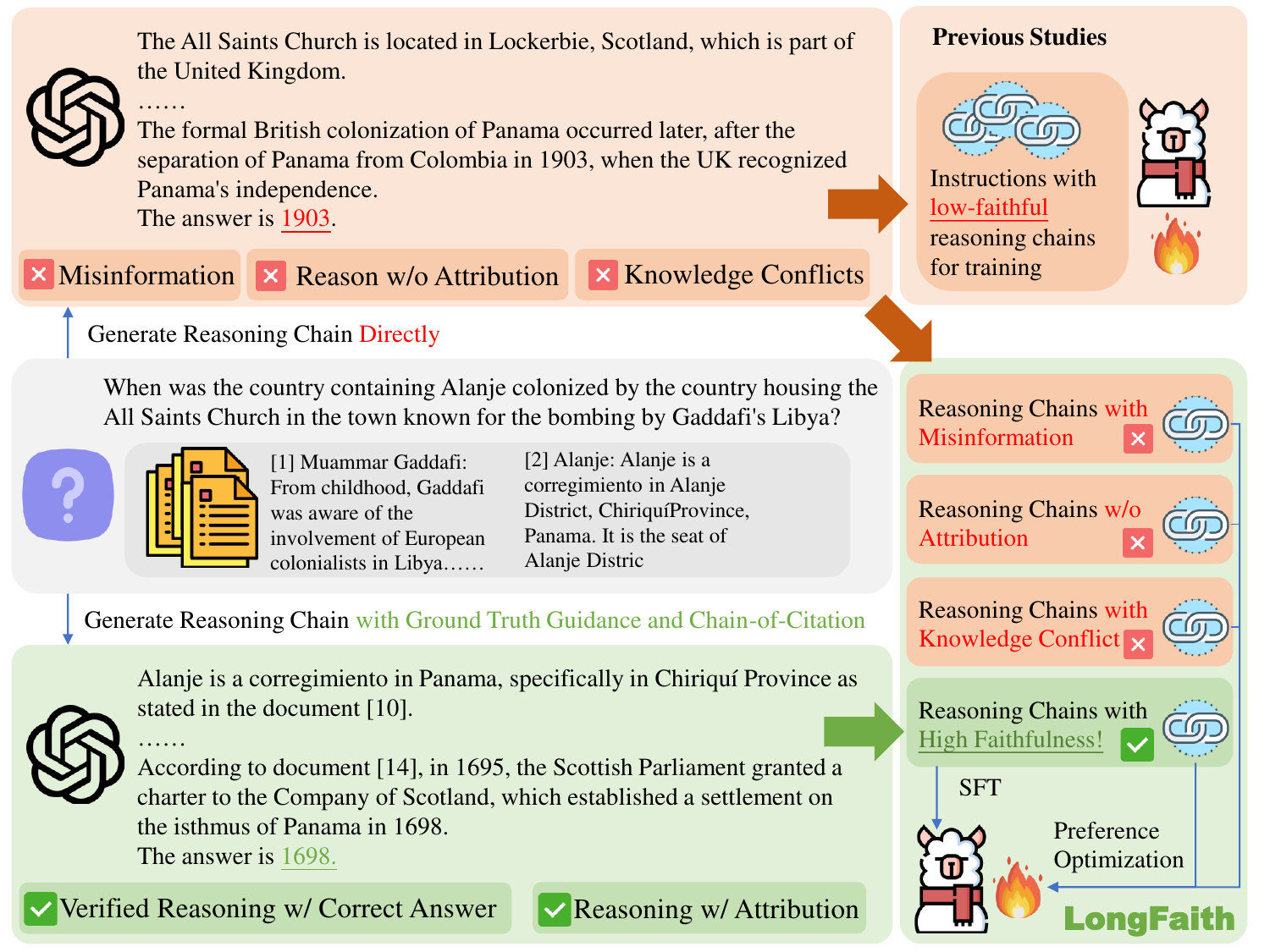}}
    \caption{Overview of \textsc{LongFaith} pipeline for synthesizing faithful long-context reasoning instruction and preference datasets. Comparing generated reasoning chains with misinformation, lack of attribution, and knowledge conflicts, \textsc{LongFaith} generates ground truth guidance prompting by chain-of-citation to build \textsc{LongFaith}-SFT. Fine-grained faithfulness is modeled by optimization on our preference datasets \textsc{LongFaith}-PO.}
    \label{fig:main}
\end{figure*}

Despite the improvements in downstream QA performance enabled by synthetic long-context reasoning instructions, concerns remain regarding the faithfulness of such generated data. Specifically: (1) \textbf{Misinformation due to lack of verification}: existing methods often generate QA pairs without rigorous rule-based verification. For instance, \cite{longmit, longreward, sealong} directly synthesize QA pairs using LLMs while bypassing verification, whereas \cite{longreward} relies on AI-generated feedback in soft dimensions rather than human annotation. (2) \textbf{Reasoning without attribution}: prompting LLMs to generate responses with citation, such as using \textit{Chain-of-Citation (CoC)} prompting~\cite{attribution, coc, learning, finegrained, cotar, alce} can enhance the credibility and interpretability of model outputs under long-context QA tasks~\cite{alce, longcite}, yet most prior works ignore to incorporate this technique during their synthesis of training instruction pairs. (3) \textbf{Potential knowledge conflicts}: some approaches \cite{longalign, longreward, longmit} over-rely on the \textit{Self-Instruct} technique~\cite{selfinstruct} to generate QA pairs, encouraging models to rely on parametric knowledge rather than grounding reasoning in explicit contextual evidence~\cite{knowledgeconflicts}. Additionally, ~\cite{longreward} feeds the query and response to a short-context reward model ignoring the long context to score, purely relying on the parametric knowledge inside LLM. These limitations underscore the necessity for a more robust pipeline that ensures the faithfulness of long-context instructions synthesis.

We propose \textbf{\textsc{LongFaith}}, a novel pipeline for synthesizing faithful long-context reasoning instruction datasets. We incorporate ground truth directly into the prompt for synthesizing long-context reasoning chains, which comprise supporting facts and the correct answer, and prompt LLMs to reason with attributions. This method ensures the faithfulness of synthesized reasoning chains without requiring costly verification by a curated rule-based evaluator, LLM-as-a-judge~\cite{llmasajudge} or human annotator. We open-source \textbf{\textsc{LongFaith}-SFT}, synthesized under the guidance of ground truth and CoC prompting. We leverage the faithful long-context reasoning chains with attributions for training, leading to performance improvements after fine-tuning \text{\llama}. Additionally, we synthesize preference datasets by sampling preference pairs around fine-grained faithfulness: (1) encouraging model to reason with attributions; (2) encouraging model to learn on verified reasoning chains; and (3) encouraging model to reason with contextual documents grounded. We open-source \textbf{\textsc{LongFaith}-PO}, synthesized by various LLMs in different sizes, which integrates all three faithfulness dimensions for preference optimization. We leverage these faithful preferred instruction pairs for training \llama, achieving performance improvements on the multi-hop reasoning dataset and LongBench~\cite{longbench}.

Our main contributions are as follows: (1) We introduce \textsc{LongFaith}, a novel pipeline for synthesizing faithful long-context reasoning instruction data. (2) We open-source \textsc{LongFaith}-SFT and \textsc{LongFaith}-PO, two synthesized datasets that are systematically designed considering multiple dimensions of faithfulness. (3) We conduct extensive experiments on two types of datasets (comprising eight sub-tasks) to show that models trained on \textsc{LongFaith} datasets can improve in long-context reasoning and QA tasks. (4) Our ablation studies illustrate the scaling potential and adaptability of \textsc{LongFaith} pipeline, underscoring its broad applicability in the development of long-context LLMs.

%% file: related.tex
\section{Related Work}
\label{sec:related}

\paragraph{Long-Context Utilization.} Amounts of studies focus on enhancing LLMs to better utilize long-context information. Model-centric approaches, for instance, optimizations on attention mechanism aim to capture specific sequential features~\cite{longformer, longnet, longlora, lminfinite}, while positional interpolation techniques are utilized to scale positional encoding while ensuring valid index ranges~\cite{pose, extending, longrope, yarn, longformer}. In addition, data-driven approaches also gain popularity, emphasizing high-quality data synthesis for fine-tuning to improve LLMs' long-context processing capabilities. For example,~\cite{effective, prolong} employ long-sequence continuous pre-training on foundation models, while~\cite{dataengineering} explores the impact of pre-training data composition and balance. Additionally, works on SFT with synthetic instructions~\cite{in2, longalign, coc, longmit} not only consider long-context understanding but also strengthen multi-hop reasoning capabilities. Lastly, preference optimization approaches~\cite{longreward, sealong} generate fine-grained pairwise preference instruction sets and incorporate training techniques~\cite{dpo, orpo}. From the perspective of improving the faithfulness of synthetic data, our work effectively addresses the shortcomings of prior studies in this area.

\paragraph{Faithful Reasoning.} Hallucination in LLMs presents a major challenge in knowledge-intensive tasks~\cite{siren, hallu}. Recent work has focused on enhancing faithful reasoning, where the goal is to trace the LLM's generated content back to reliable sources and ensure its factual grounding. ~\cite{cotar, attribution, coc} aim to improve the identification and verification of attributions by focusing on generating reasoning outputs that link claims to specific sources. Benchmarks such as ~\cite{alce, automatic} evaluate the quality of citations and highlight the limitations of current systems in providing citation support to ensure more reliable output. Additionally, integrating external knowledge sources has gained attention, which use retrieval-augmented generation (RAG) methods to facilitate deep and faithful reasoning~\cite{tog, tog2}. Our \textsc{LongFaith} is motivated by previous work, towards faithful reasoning in long-context reasoning tasks.

%% file: method.tex
\section{LongFaith}
\label{sec:method}

In this section, we present an exposition of \textsc{LongFaith} pipeline. Specifically, we explain how it synthesizes \textsc{LongFaith}-SFT for supervised fine-tuning and \textsc{LongFaith}-PO for preference optimization from the perspective of faithfulness.

\paragraph{Synthesize Reasoning Chains with High Faithfulness.} Previous studies~\cite{longalign, longlora, longmit, longreward, sealong} tend to directly distill synthesized long-context QA and reasoning instructions for training without filtering out incorrect information. These low-faithfulness synthesized data limit the performance improvements of the trained models. In response to this challenge, \textsc{LongFaith} integrates ground truth into the synthesized reasoning chains. For a sample \(S\) from the training set \(S = (Q, D, A, F)\), where \(Q\) is the reasoning question, \(D\) is the full document used for querying, \(A\) is the correct answer, and \(F\) represents the supporting facts where \(F \in D\). We use the LLM \(M_{\text{syn}}\) to synthesize the reasoning chain as follows:

\begin{equation}
    O_c = M_{\text{syn}}(P_{\text{coc}}, Q, F, A)
\end{equation}

Here, \(O_c\) represents the output of \(M_{\text{syn}}\), which is a step-by-step reasoning chain. The prompt \(P_{\text{coc}}\) utilizes a chain-of-citation~\cite{coc} prompting approach, requiring the model to reason with attribution (e.g., "Let's reason step by step while citing the document using [1], [2], etc."). The prompt template is shown in Figure~\ref{fig:pt_1}.

\paragraph{\textsc{LongFaith}-SFT Dataset.}

Towards training a downstream LLM to reason with high faithfulness for a long-context QA task, we construct the dataset for supervised fine-tuning, where each instruction pair is built as follow:

\begin{equation}
    I_{\text{sft}} = \{\text{input} = (P_{\text{coc}}, Q, D), \text{output} = O_c\}
\end{equation}

\paragraph{Synthesize Reasoning Chain with Questionable Faithfulness.}

To model fine-grained preferences, we address three challenges that affect the faithfulness of synthesized instructions: (1) misinformation due to lack of verification, (2) reasoning without attribution, and (3) potential knowledge conflicts. We synthesize reasoning chains with questionable faithfulness, including \textbf{reasoning chains with misinformation} as follows:

\begin{equation}
    O_m = M_{\text{syn}}(P_{\text{coc}}, Q, D)
\end{equation}

Since there is no ground truth to guide the synthesis, the output \(O_m\) may contain errors in reasoning, as illustrated in Figure~\ref{fig:pt_2}, where the model generates an incorrect answer of "1903" instead of the correct answer "1698". This hallucination is common in synthesized data from previous works unless rules or human experts are involved in filtering~\cite{coc}. Next, we synthesize \textbf{reasoning chains without attribution}:

\begin{equation}
    O_{\text{cot}} = M_{\text{syn}}(P_{\text{cot}}, Q, F, A)
\end{equation}

Here, the CoT~\cite{cot} prompting only requires the model to provide step-level reasoning, but as shown in Figure~\ref{fig:pt_3}, reasoning without attribution not only loses interpretability and credibility~\cite{alce, attribution}, but our results in Tab.~\ref{tab:ab_models} (Sec.~\ref{sec:exp}) also demonstrate that CoT prompting performs worse than CoC. Finally, we synthesize \textbf{reasoning chains with potential knowledge conflicts}:

\begin{equation}
    O_{\text{kc}} = M_{\text{syn}}(P_{\text{cot}}, Q, A)
\end{equation}

Since no context is provided, the model relies solely on its parametric knowledge for reasoning, as shown in Figure~\ref{fig:pt_4}, where the model states, "Panama was not colonized by the United Kingdom; Panama was colonized by Spain," based on internal parametric knowledge rather than the contextual documents. Previous studies~\cite{longreward} using short-context reward models observes performance degradation by ignoring long-context materials, highlighting the limitation of knowledge conflicts in affecting LLM's performance in long-context QA and reasoning tasks.

\begin{table}[!t]
    \tabcolsep 1.5pt
    \centering
    \small
    \begin{tabular}{lccccc}
        \toprule
        \multicolumn{6}{c}{\textit{Synthesis of Reasoning Chains}} \\
        \midrule
        \multicolumn{1}{c}{\textbf{Models}} & \textbf{Prompt} & \textbf{w/ GT} & \textbf{w/ Doc} & {\textbf{Output}} & \textbf{Size} \\
        \midrule
        \textit{Q-7B} & CoC & \checkmark & \checkmark & 1 & 1-8K \\
        \textit{Q-7B} & CoT & \checkmark & \checkmark & 2 & 1-8K \\ 
        \textit{Q-7B} & CoC & \ding{55} & \checkmark & 3 & 1-8K \\ 
        \textit{Q-7B} & CoT & \checkmark & \ding{55}  & 4 & 1-8K\\
        \textit{L8,L70,G} & CoC & \checkmark & \checkmark & 5& 2K  \\
        \textit{L8,L70,G} & CoT & \checkmark & \checkmark & 6 & 2K \\ 
        \textit{L8,L70,G} & CoC & \ding{55} & \checkmark & 7 & 2K \\ 
        \textit{L8,L70,G} & CoT & \checkmark & \ding{55} & 8 & 2K \\  
        \midrule
        \multicolumn{6}{c}{\textit{Datasets for Supervised Fine-tuning}} \\
        \midrule
        \multicolumn{1}{c}{\textbf{Name}} & \textbf{Models} & \textbf{Instruction} & \textbf{Output} & & \textbf{Size} \\
        \midrule
        \textbf{LF-SFT} & \textit{Q-7B} & CoC & 1 & & 1-8K \\
        \;\;w/o CoC & \textit{Q-7B} & CoT & 2 & & 1-8K \\
        \;\;w/o GT & \textit{Q-7B} & CoC & 3 & & 1-8K \\
        \;\;w/o Doc & \textit{Q-7B} & CoC & 4 & & 1-8K \\
        \textbf{LF-SFT} & \textit{L8,L70,G} & CoC & 5 & & 2K \\
        \;\;w/o CoC & \textit{L8,L70,G} & CoT & 6 & & 2K \\
        \;\;w/o GT & \textit{L8,L70,G} & CoC & 7 & & 2K \\
        \;\;w/o Doc & \textit{L8,L70,G} & CoC & 8 & & 2K \\
        \midrule
        \multicolumn{6}{c}{\textit{Datasets for Preference Optimization}} \\
        \midrule
        \multicolumn{1}{c}{\textbf{Name}} & \textbf{Models} & \textbf{Instruction} & \textbf{Chosen} & \textbf{Rejected} & \textbf{Size}\\
        \midrule
        \;\;w/ CoC & \textit{Q-7B} & CoC & 1 & 2 & 1-8K \\
        \;\;w/ GT & \textit{Q-7B} & CoC & 1 & 3 & 1-8K \\
        \;\;w/ Doc & \textit{Q-7B} & CoC & 1 & 4 & 1-8K \\
        \textbf{LF-PO} & \textit{Q-7B} & CoC & 1 & 2,3,4 & 1-8K \\
        \;\;w/ CoC & \textit{L8,L70,G} & CoC & 5 & 6 & 2K \\
        \;\;w/ GT & \textit{L8,L70,G} & CoC & 5 & 7 & 2K \\
        \;\;w/ Doc & \textit{L8,L70,G} & CoC & 5 & 8 & 2K \\
        \textbf{LF-PO} & \textit{L8,L70,G} & CoC & 5 & 6,7,8 & 2K \\
        \bottomrule
    \end{tabular}
    \caption{Statistics of synthesized datasets for SFT and PO. We first synthesize large-scale reasoning chains and then refactor them to datasets, where the second stage does not require llm inference. \textbf{\textit{Q-7B}} means \textit{\qwen}, \textbf{\textit{L8}} means \textit{\llama}, \textbf{\textit{L70}} means \textit{\llamal} and \textbf{\textit{G}} means \textit{\gpt}. \textbf{GT} means Ground Truth, \textbf{CoC} means chain-of-citation, \textbf{Doc} means contextual documents, and \textbf{LF} means \textsc{LongFaith}. 1-8K includes \{1K, 2K, 4K, 8K\}.}
    \label{tab:synset_stat}
\end{table}

\begin{table*}[t]
\centering
\scriptsize
\tabcolsep 3.0pt
\begin{tabular}{lcccccccccccccccc}
\toprule
\multirow{2}{*}{\textsc{\llama}} &
\multicolumn{2}{c}{\textbf{\musique}} &
\multicolumn{2}{c}{\textbf{2Wiki}} &
\multicolumn{2}{c}{\textbf{HotpotQA}} &
\multicolumn{2}{c}{\textbf{Qasper(S)}} &
\multicolumn{2}{c}{\textbf{MFQA-En(S)}} &
\multicolumn{2}{c}{\textbf{\musique(M)}} &
\multicolumn{2}{c}{\textbf{2Wiki(M)}} &
\multicolumn{2}{c}{\textbf{\hotpot(M)}} \\
\cmidrule(r){2-3} \cmidrule(r){4-5} \cmidrule(r){6-7} \cmidrule(r){8-9} \cmidrule(r){10-11} \cmidrule(r){12-13} \cmidrule(r){14-15} \cmidrule(r){16-17}
& F1 & SubEM & F1 & SubEM & F1 & SubEM & F1 & SubEM & F1 & SubEM & F1 & SubEM & F1 & SubEM & F1 & SubEM  \\
\midrule
\multicolumn{17}{c}{\textit{Zero-Shot Prompting}} \\
\midrule
\quad + CoT & 15.9 & 56.8 & 34.0 & 83.8 & 20.8 & 78.6 & 3.2 & 22.0 & 5.7 & 29.3 & 14.1 & 43.5 & 30.1 & 77.0 & 13.4 & 60.5 \\
\quad + CoC & 25.8 & 64.2 & 43.6 & 86.2 & 32.7 & 76.6 & 4.6 & 26.0 & 7.0 & 32.7 & 11.8 & 41.0 & 28.1 & 79.5 & 19.9 & 58.0 \\
\midrule
\multicolumn{17}{c}{\textit{Superivised Fine-tuning}} \\
\midrule
\quad + LongAlpaca & 21.6 & 50.2 & 47.8 & 80.4 & 32.7 & 76.6 & 5.7 & 25.0 & 5.8 & 30.7 & 8.5 & 48.5 & 25.4 & 77.0 & 12.5 & 61.0 \\
\quad + LongAlign & 24.8 & 48.4 & 55.6 & 84.2 & 51.0 & 79.2 & 6.5 & 24.0 & 10.7 & \textbf{38.7} & 15.0 & 40.0 & 33.4 & 76.5 & 35.8 & 61.0 \\
\quad + \musique-Attribute & 13.9 & 19.2 & 23.9 & 49.6 & 20.2 & 37.2 & 10.0 & 11.5 & 8.3 & 12.0 & 15.2 & 26.5 & 21.2 & 55.0 & 25.6 & 41.0 \\
\quad + LongMIT & 4.9 & 33.0 & 3.3 & 58.0 & 10.1 & 63.6 & 9.5 & 18.5 & 5.6 & 30.0 & 7.5 & 29.0 & 3.6 & 55.5 & 23.7 & 50.0 \\
\quad + LongReward-SFT & 6.2 & 48.4 & 23.3 & 80.0 & 15.6 & 74.2 & 2.6 & 22.5 & 0.5 & \underline{34.0} & 1.1 & 43.0 & 6.6 & 71.5 & 8.9 & 54.0 \\
\quad + \textsc{SeaLong}-SFT & 31.3 & \underline{64.6} & 55.8 & \underline{89.2} & 59.4 & \textbf{83.0} & 14.5 & 26.0 & 18.6 & 31.3 & 24.1 & \textbf{59.5} & 34.1 & \underline{84.5} & 37.3 & \underline{69.0} \\
\quad + \textsc{LongFaith}-SFT & \underline{56.8} & 62.8 & \textbf{73.8} & 85.6 & \textbf{70.5} & 80.8 & \underline{36.9} & \underline{29.5} & \textbf{47.0} & 32.0 & \underline{50.1} & \underline{56.5} & \underline{63.9} & 82.0 & \underline{53.1} & 68.0 \\
\midrule
\multicolumn{17}{c}{\textit{Preference Optimization}} \\
\midrule
\quad + \textsc{LongReward}-PO & 3.3 & 46.0 & 14.3 & 76.6 & 8.9 & 71.2 & 1.6 & 21.0 & 0.1 & 32.7 & 0.0 & 37.5 & 4.4 & 67.0 & 3.3 & 53.0 \\
\quad + \textsc{SeaLong}-PO & 30.2 & 60.4 & 50.1 & \textbf{89.4} & 58.3 & \textbf{83.4} & 17.1 & 28.0 & 20.1 & 32.0 & 18.1 & 53.3 & 34.0 & \textbf{86.0} & 40.2 & \textbf{69.5} \\
\quad + \textsc{LongFaith}-PO & \textbf{60.5} & \textbf{66.4} & \underline{68.0} & 85.0 & \underline{65.4} & 81.2 & \textbf{38.1} & \textbf{30.5} & \underline{46.7} & 32.0 & \textbf{50.2} & 52.0 & \textbf{73.7} & 83.5 & \textbf{55.6} & 67.5 \\
\midrule
\end{tabular}
\caption{Main experiment results on three multi-hop reasoning test sets and five long-context QA test sets from LongBench. The best results are in \textbf{bold} and second-best are \underline{underlined}. \textbf{(S)} means single-doc QA task and \textbf{(M)} means multiple-docs QA task in LongBench. \textsc{LongFaith}-SFT and \textsc{LongFaith}-PO are synthesized by \textit{\gpt} both in 2K size. To ensure fairness, we sample first 2K examples from baseline datasets.}
\label{tab:longbench_exp}
\end{table*}

\paragraph{\textsc{LongFaith-PO} Dataset.}

Towards training a downstream LLM to address three challenges above in long-context reasoning, we force the LLM to learn reasoning with high faithfulness while rejecting outputs of questionable faithfulness:

\begin{equation}
\begin{split}
    I_{\text{po}} = \{\text{input} = (P_{\text{coc}}, Q, D), \\
    \text{chosen} = O_c, \text{rejected} = O_r\}
\end{split}
\end{equation}

where \(O_r\) is a combination of (\(O_m\), \(O_{\text{cot}}\), \(O_{\text{kc}}\)), or a subset of them.

%% file: exp.tex
\section{Experiments}
\label{sec:exp}

\begin{table*}[t]
\centering
\small
\tabcolsep 4.0pt
\begin{tabular}{lcccccccccc}
\toprule
\multirow{2}{*}{\textsc{\llama}} &
\multicolumn{4}{c}{\textbf{\musique}} &
\multicolumn{3}{c}{\textbf{\twowiki}} &
\multicolumn{3}{c}{\textbf{\hotpot}} \\
\cmidrule(r){2-5} \cmidrule(r){6-8} \cmidrule(r){9-11}
& Overall & 2-Hop & 3-Hop & 4-Hop & Overall & 2-Hop & 4-Hop & Overall & Bridge & Comparison\\
\midrule
\multicolumn{11}{c}{\textit{Zero-Shot Prompting}} \\
\midrule
\quad + CoT & 11.0 & 7.5 & 16.2 & 12.0 & 29.0 & 22.3 & 54.3 & 17.4 & 17.5 & 17.0 \\
\quad + CoC & 19.0 & 16.1 & 22.7 & 20.7 & 39.2 & 31.4 & 68.6 & 30.4 & 28.6 & 38.6 \\
\midrule
\multicolumn{11}{c}{\textit{Superivised Fine-tuning}} \\
\midrule
\quad + \textsc{LongFaith}-SFT & 40.6 & 44.1 & 37.7 & 35.9 & 55.4 & 51.1 & 71.4 & 53.6 & 57.0 & 37.5 \\
\quad\quad w/o CoC & 40.2 & 41.7 & 39.6 & 37.0 & 51.8 & 48.9 & 62.9 & 52.0 & 56.6 & 30.7 \\
\quad\quad w/o GT & 30.4 & 31.9 & 28.6 & 29.3 & 55.8 & 49.6 & 79.0 & 56.6 & 54.9 & \textbf{64.8} \\
\quad\quad w/o Doc & 20.0 & 23.6 & 18.2 & 13.0 & 55.8 & 47.1 & \textbf{88.6} & 47.4 & 45.9 & 54.5 \\
\midrule
\multicolumn{11}{c}{\textit{Preference Optimization}} \\
\midrule
\quad\quad w/ GT-PO & 44.0 & 45.7 & 42.9 & 41.3 & 56.0 & 50.4 & 77.1 & 54.4 & 58.3 & 36.4 \\
\quad\quad w/ CoC-PO & 43.6 & 44.5 & 44.8 & 39.1 & 53.2 & 48.6 & 70.5 & 56.2 & 59.2 & 42.0 \\
\quad\quad w/ Doc-PO & 41.4 & 42.5 & 40.3 & 40.2 & 56.0 & 52.7 & 68.6 & 56.4 & 59.5 & 42.0 \\
\quad + \textsc{LongFaith}-PO & \textbf{46.6} & \textbf{47.2} & \textbf{48.1} & \textbf{42.4} & \textbf{59.0} & \textbf{55.9} & 70.5 & \textbf{58.6} & \textbf{59.7} & 53.4 \\
\midrule
\end{tabular}
\caption{Main experiment results on three long-context multi-hop reasoning datasets using the Exact-Match(\textbf{EM}) metric. The best results are in \textbf{bold}. The training set has 2K samples for both SFT and PO, synthesized by \textit{\qwen}. Detail statistics of synthetics datasets are presented in Tab.~\ref{tab:synset_stat}.}
\label{tab:multihop_exp}
\end{table*}

\subsection{Implementation Details} Following previous studies, we leverage the training set of \musique~\cite{musique}, which is build on Wikipedia documents with supporting documents and correct answers. The officially retrieved 20 documents are provided and read only once in the input context in distractor setting. The statistics of training set is given in Tab.~\ref{tab:train_stat}, covering \textit{1K}, \textit{2K}, \textit{4K} and \textit{8K}, where the balance of questions with different hops are considered. Following the pipeline we describe in Sec.~\ref{sec:method}, reasoning chains are samples to build \textsc{LongFaith}-SFT and \textsc{LongFaith}-PO. The statistics are presented in Tab.~\ref{tab:synset_stat}.

We conduct our experiments on a Linux server equipped with 4 A100-SXM4-40GB GPUs. For data synthesis of long-context reasoning instructions, we take \textit{\llama}~\cite{llama3}, \textit{\qwen}~\cite{qwen2.5}, \textit{\llamal} and \textit{\gpt}~\cite{gpt-4o} as generators and prompt LLMs to synthesize reasoning chains with vLLM~\cite{vllm}. We adopt the LoRA technique~\cite{lora} for fine-tuning and ORPO technique~\cite{orpo} for preference optimization using the LLaMA-Factory framework~\cite{llamafactory} to train \textit{\llama}. Hyperparameters of post-training are given in App.~\ref{sec:hyperparameters}.

\begin{figure}[!t]
    \centering
    \centerline{\includegraphics[width=\linewidth]{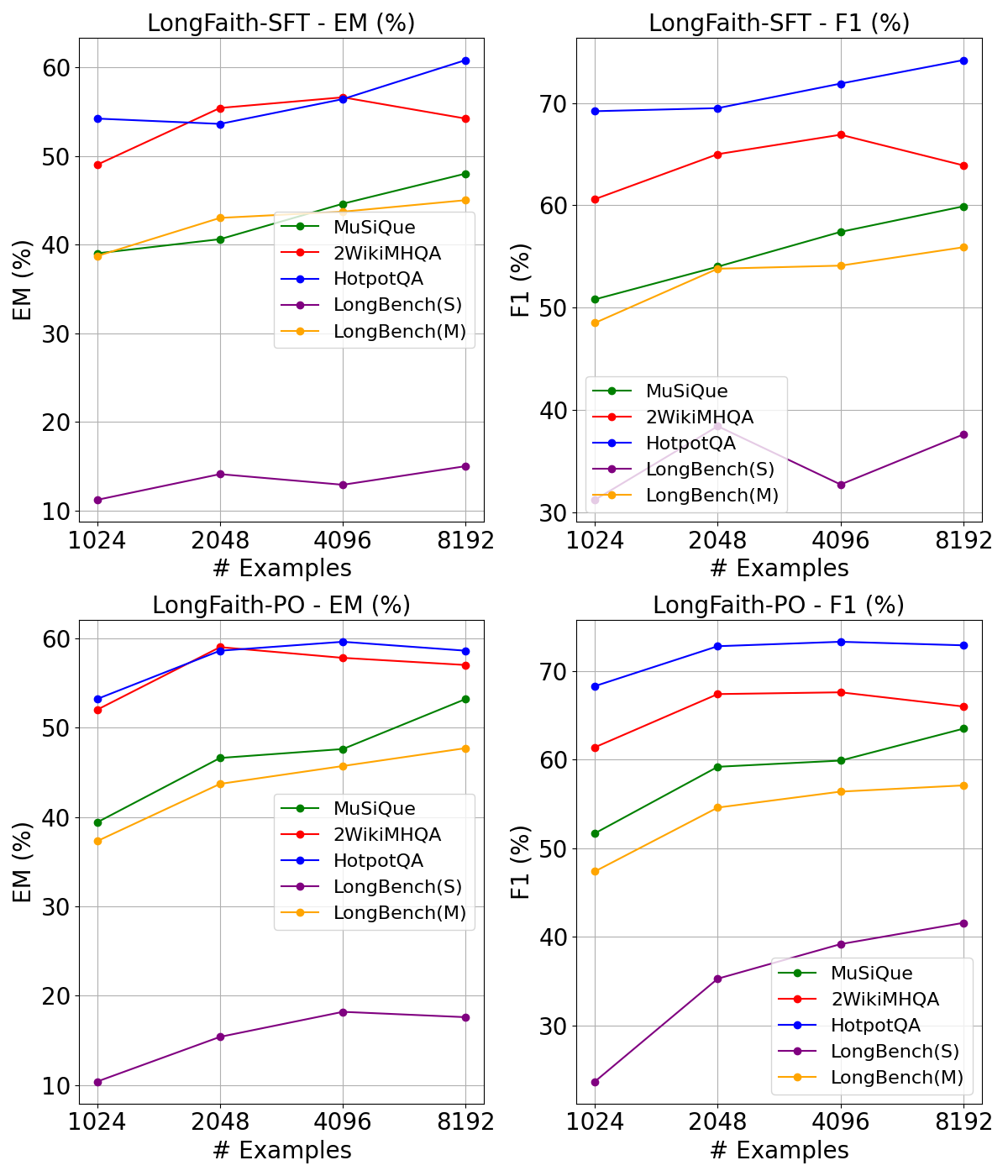}}
    \caption{Performance of \textit{\llama} trained on different size of instructions synthesized by \textit{\qwen} from \textit{1K} to \textit{8K}, evaluated by \textbf{EM} and \textbf{F1} metrics on multi-hop reasoning sets and LongBench.}
    \label{fig:scale}
\end{figure}

\subsection{Evaluation Setup}
Following prior work~\cite{coc}, we utilize \textbf{three multi-hop reasoning datasets}, including \musique~\cite{musique}, \twowiki~\cite{twowiki}, and \hotpot~\cite{hotpotqa}, evaluating in distractor-setting, where the officially retrieved 10 or 20 documents are provided and read only once in the input context. We adopt the test sets sampled by~\cite{multihopdatasets}, with 500 examples in each set. Furthermore, in line with previous studies~\cite{longmit, longreward, sealong}, we assess the performance on \textbf{LongBench}~\cite{longbench}, which includes two test sets for single-doc QA including Qasper~\cite{qasper} and MultiFieldQA-EN~\cite{longbench}, as well as three test sets for multi-docs QA tasks including \hotpot, \twowiki, and \musique. Notably, although there is \textbf{an overlap in multi-hop reasoning tasks}, the LongBench version \textbf{further extends the lengths of document text}. To apply CoC prompting, single document is split into 20 even paragraphs with order. The statistics of datasets are listed in Tab.~\ref{tab:dataset_stat}. 

To ensure fairness, we use Substring Exact-Match (\textbf{SubEM})~\cite{helmet, sealong} metric in main experiments, in case that models trained on baseline datasets are not good at instructions following to summarize the answer with "The answer is", and SubEM goes through the whole response to check whether the answer is in. Furthermore, following previous work~\cite{quac, retrieval, coc}, we use \textbf{EM} metric and \textbf{F1} scores for the trimmed part after "The answer is" for evaluation in main experiments and ablation studies. Comparing with LLM-as-a-Judge~\cite{longalign, longmit, longreward} using strong API models like GPT-4o, the rule-based metrics cost much lower.

\subsection{Baselines}
We compete \textsc{LongFaith}-SFT and \textsc{LongFaith}-PO with datasets proposed in previous studies, including \textsc{LongAlpaca}~\cite{longlora}, \textsc{LongAlign}~\cite{longalign}, \textsc{MuSiQue-Attribute}~\cite{coc}, \textsc{LongMIT}~\cite{longmit}, \textsc{LongReward}-SFT~\cite{longreward}, \textsc{SeaLong}-SFT~\cite{sealong} for supervised fine-tuning, and \textsc{LongReward}-PO and \textsc{SeaLong}-PO for preference optimization. All of them aim at enhancing the performance of LLMs on long-context understanding, reasoning, and QA tasks. To ensure fairness, we keep the training setting consistent with App.~\ref{sec:hyperparameters} and truncate the size of training samples to a maximum of 2K.

\begin{figure*}[t]
    \centering
    \centerline{\includegraphics[width=\linewidth]{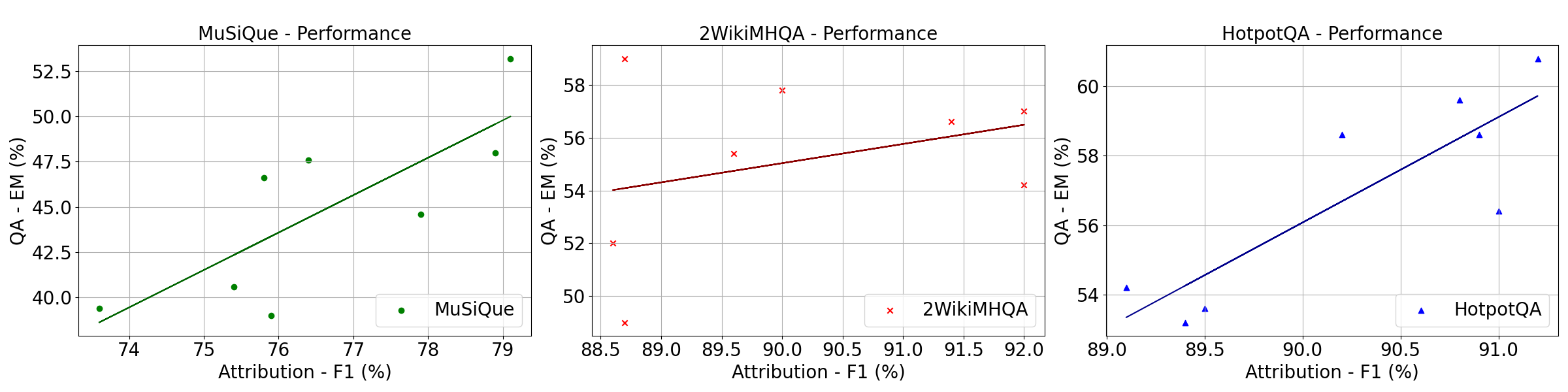}}
    \caption{Scatter plot with a linear regression line fitting the relationship between \textbf{QA - EM} and \textbf{Attribution - F1} metrics on three long-context multi-hop reasoning test sets. A point refers to the performance of a model trained with a specific size between \textit{1K} to \textit{8K} by SFT or PO.}
    \label{fig:point}
    \vspace{-5pt}
\end{figure*}

\subsection{Main Results}  
\paragraph{\textsc{LongFaith} Outperforms Previous Datasets.} Following previous work and to ensure a fair comparison, we evaluate the performance of \textsc{LongFaith} on multi-hop reasoning test sets~\cite{musique, twowiki, hotpotqa} and LongBench~\cite{longbench}, comparing it against baseline methods, including zero-shot prompting with \llama and models trained on synthetic datasets from previous works. As shown in Tab.~\ref{tab:longbench_exp}, \textsc{LongFaith} outperforms baseline datasets on most test sets. The performance of the model trained on \textsc{LongFaith}-PO surpasses that trained on \textsc{LongFaith}-SFT. This aligns with our expectations: compared to directly using positive samples for supervised fine-tuning, incorporating rejected samples to provide more fine-grained faithfulness preferences for optimization leads to better improvements in long-context reasoning and QA capabilities. We observe that some synthetic instruction sets degrade performance compared to native \llama. This proves that datasets with questionable faithfulness are even harmful to long-context reasoning ability of LLMs.

\paragraph{\textsc{LongFaith} Arrives at the Correct Answer without Redundant Exploration.} We find that on \twowiki, \hotpot, and part of tasks in LongBench, \textsc{SeaLong} achieves a slight advantage in the SubEM metric against \textsc{LongFaith}, but fails in F1 scores. We investigate the length of response and present in Tab.~\ref{tab:output_stat}. It turns out that the LLM trained on \textsc{SeaLong} conducts redundant exploration in response, producing more noisy content related to the answer, but actually arrives at a wrong answer, which means SubEM metric is easily to be "hacked". In contradiction, F1 scores requires to truncate the part after "The answer is", which demonstrates that a model trained on \textsc{LongFaith} datasets can arrive at the correct answer without redundant exploration and achieve a high score in a more strict metric. A case study is shown in Fig.~\ref{fig:hack} in Appendix.

\paragraph{Generalization.} Based on statistics from Tab.~\ref{tab:baseline_stat}, the main experiment demonstrates that \textsc{LongFaith} uses instructions with shorter context as input compared to baseline methods, reducing training costs while generalizing to LongBench tasks that require processing an average of 24K-70K tokens as input. This further highlights the generalization ability of our pipeline.

\begin{table*}[t]
\centering
\small
\tabcolsep 7.0pt
\begin{tabular}{lcccccccccc}
\toprule
\multicolumn{1}{c}{\textsc{\llama}} &
\multicolumn{2}{c}{\textbf{\musique}} &
\multicolumn{2}{c}{\textbf{2WikiMHQA}} &
\multicolumn{2}{c}{\textbf{\hotpot}} &
\multicolumn{2}{c}{\textbf{LongBench(S)}} &
\multicolumn{2}{c}{\textbf{LongBench(M)}} \\
\cmidrule(r){2-3} \cmidrule(r){4-5} \cmidrule(r){6-7} \cmidrule(r){8-9} \cmidrule(r){10-11}
\quad + \textsc{LongFaith} & EM & F1 & EM & F1 & EM & F1 & EM & F1 & EM & F1 \\
\midrule
\multicolumn{11}{c}{\textit{Superivised Fine-tuning}} \\
\midrule
\quad w/ \textit{\llama} & 35.4 & 48.4 & 59.4 & 69.5 & 54.6 & 67.7 & 9.9 & 29.0 & 42.2 & 48.4 \\
\quad w/ \textit{\qwen} & 40.6 & 54.0 & 55.4 & 65.0 & 53.6 & 69.5 & 14.1 & 38.4 & 43.0 & 53.8 \\
\quad w/ \textit{\llamal} & \textbf{44.8} & \textbf{58.1} & 54.0 & 64.4 & 54.4 & 69.5 & 16.0 & 41.0 & 44.8 & \textbf{56.7} \\
\quad w/ \textit{\gpt} & 41.6 & 56.8 & \textbf{64.6} & \textbf{73.8} & \textbf{55.4} & \textbf{70.5} & \textbf{16.9} & \textbf{42.0} & \textbf{47.2} & 55.7 \\
\midrule
\multicolumn{11}{c}{\textit{Preference Optimization}} \\
\midrule
\quad w/ \textit{\llama} & 41.2 & 53.2 & 57.4 & 67.0 & 55.6 & 68.5 & 14.7 & 36.9 & 44.0 & 55.3 \\
\quad w/ \textit{\qwen} & 46.6 & 59.2 & 59.0 & 67.4 & \textbf{58.6} & \textbf{72.8} & 15.4 & 35.3 & 44.8 & 55.6 \\
\quad w/ \textit{\llamal} & \textbf{50.4} & \textbf{63.2} & 52.8 & 62.7 & 57.2 & 71.0 & \textbf{16.4} & 40.0 & \textbf{48.3} & 59.4 \\
\quad w/ \textit{\gpt} & 48.4 & 60.5 & \textbf{59.8} & \textbf{68.0} & 49.8 & 65.4 & 15.9 & \textbf{42.4} & 45.0 & \textbf{59.8} \\
\midrule
\end{tabular}
\caption{Ablation study on various LLMs for synthesizing \textsc{LongFaith}-SFT and \textsc{LongFaith}-PO in the size of 2K. The base model for training and testing is \textit{\llama}. \textbf{(S)} and \textbf{(M)} refer to Single-doc QA and Multi-docs QA in LongBench.}
\label{tab:ab_models}
\end{table*}

\begin{figure}[t]
    \centering
    \centerline{\includegraphics[width=\linewidth]{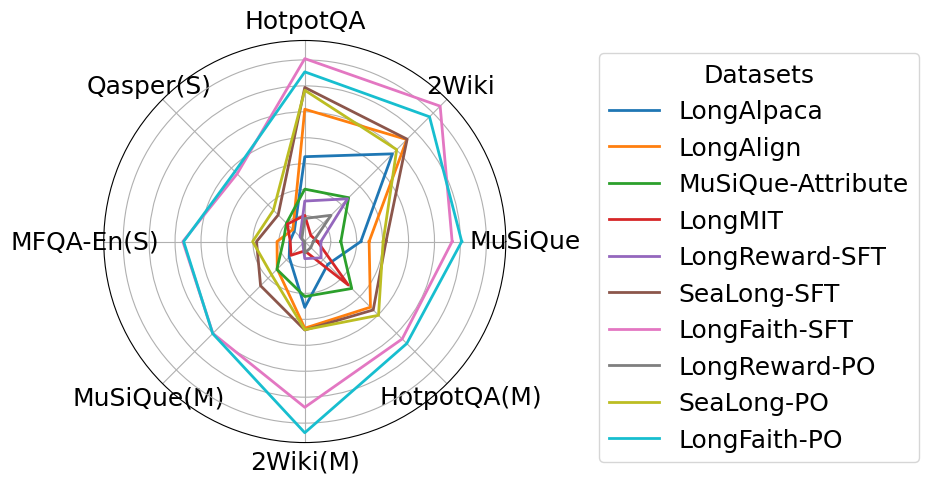}}
    \caption{Visualization of F1 scores in Tab.~\ref{tab:longbench_exp}.}
    \label{fig:radar}
\end{figure}

\subsection{Analysis}

\paragraph{Exploration on Different Perspective of Faithfulness.} To validate the specific impact of different dimensions of faithfulness, we fine-tune models using negative samples as output and optimize using preference datasets that reject only a subset of negative samples. The statistics of the constructed datasets are shown in Tab.~\ref{tab:synset_stat}. Since each task in LongBench contains no more than 200 questions, performance evaluations can be prone to errors, so we chose to test on multi-hop reasoning datasets. Experimental results are shown in~\ref{tab:multihop_exp}. The models trained with \textsc{LongFaith}-SFT and \textsc{LongFaith}-PO achieved high performance respectively in SFT and PO especially in F1 scores, as expected. 

However, we note that in 4-Hop part of 2Wiki and comparison part of \text{\hotpot}, \textsc{LongFaith}-SFT w/o CoC and w/o GT demonstrated better performance. Analysis reveals that for the question "Do both films, Cuban Colony and Prathyartha, have directors from the same country?", as the training set \text{\musique} used specific entities as answers, the model responds "Both directors are from the same country, which is India. The answer is India.". Actually, the correct answer is "yes." Model trained on \textsc{LongFaith} w/o GT and \textsc{LongFaith} w/o Doc performed better with more exploration, but also lost overall performance due to hallucinations. Models trained on all PO datasets outperformed those trained using only positive samples for SFT, demonstrating the performance improvement brought by each fine-grained, credible preference. Finally, models trained on \textsc{LongFaith}-PO, which integrates three dimensions of faithfulness, achieved the best overall performance.

\paragraph{Scalability and Performance Gains.} We explore the scaling-up potential of \textsc{LongFaith} on multi-hop reasoning test sets and LongBench. As presented in Tab.~\ref{tab:dataset_stat}, we train \textit{\llama} using \textsc{LongFaith}-SFT and \textsc{LongFaith}-PO synthesized by \textit{\qwen} across four dataset sizes, ranging from \textit{1K} to \textit{8K}. According to the performance trend in Fig.~\ref{fig:scale}, \textsc{LongFaith} generally exhibits scaling-up potential, indicating that expanding the training dataset can further enhance performance. Moreover, \textsc{LongFaith}-PO, which incorporates fine-grained preference optimization, demonstrates a more stable upward trend compared to \textsc{LongFaith}-SFT, particularly in LongBench tasks. This result validates the robustness of the \textsc{LongFaith} pipeline.

\paragraph{Attribution-Based Reasoning Leads to Higher Performance.} Utilizing CoC prompting for reasoning with attributions not only outperforms CoT in performance, as it presents in Tab.~\ref{tab:multihop_exp}, but also provides greater interpretability and faithfulness as shown in Fig.~\ref{fig:main}. We use Attribution F1 as a metric to quantify the model's attribution capability using annotated supporting facts. Under CoC prompting, we analyze the references within reasoning chains, matching them against supporting facts like [1], [2], etc., and compute F1 scores based on recall and precision. We evaluate the attribution capability and overall performance of \llama trained on \textsc{LongFaith}-SFT and \textsc{LongFaith}-PO across four sizes and visualize the results in a scatter plot. The findings in~\ref{fig:point} demonstrate a strong positive correlation between attribution capability and model performance, further validating the effectiveness of the \textsc{LongFaith} pipeline.

\paragraph{Impact of LLM Selection for Synthesis.} We experimented with different LLMs for synthesis, including smaller open-source LLMs such as \textit{\llama}, \textit{\qwen}, and larger open-source models like \textit{\llamal}, as well as a closed-source API model, \textit{\gpt}, to synthesize \textsc{LongFaith}-SFT and \textsc{LongFaith}-PO for training \llama. The performance test results are presented in Tab.\ref{tab:ab_models}. Using stronger closed-source API models to synthesize \textsc{LongFaith}-SFT led to a stronger performance boost, which aligns with intuition and previous work~\cite{longmit}. However, an interesting finding is that the \textsc{LongFaith}-PO synthesized with different base LLMs did not show significant performance differences in preference optimization. Even smaller model like \textit{\qwen}, are able to synthesize high-quality reasoning chains, with performance on some datasets matching or even surpassing \textit{\gpt}. This highlights the robustness of the \textsc{LongFaith} pipeline, which is capable of modeling fine-grained preferences to synthesize high-quality instructions.

\begin{table}[t]
\centering
\small
\tabcolsep 5.0pt
\begin{tabular}{lcccc}
\toprule
\multirow{2}{*}{\textsc{Llama-3.1-8B}} &
\multicolumn{2}{c}{\textbf{LongBench(S)}} &
\multicolumn{2}{c}{\textbf{LongBench(M)}} \\
\cmidrule(r){2-3} \cmidrule(r){4-5}
& EM & F1 & EM & F1 \\
\midrule
\quad w/ \textit{10 docs} & 11.2 & 31.7 & 39.7 & 52.2 \\
\quad w/ \textit{20 docs} & 15.4 & 35.3 & 44.8 & 55.6 \\
\quad w/ \textit{30 docs} & 16.0 & 36.0 & 46.1 & 56.7 \\
\quad w/ \textit{40 docs} & 16.9 & 37.8 & 47.2 & 58.1 \\
\midrule
\end{tabular}
\caption{Effectiveness of \textsc{LongFaith} on LongBench as context length increases.}
\label{tab:context_length}
\end{table}

\paragraph{Effectiveness of \textsc{LongFaith} as Context Length Increases.}  We further investigate the impact of increasing context length on the performance of models trained with the \textsc{LongFaith} pipeline. Specifically, we manipulate the number of documents included in the training set \textit{\musique} to simulate varying context lengths during training. By incrementally increasing the number of documents from 10 to 40, we assess how the model's reasoning ability scales when exposed to longer input contexts. The results, as reported in Tab.~\ref{tab:context_length}, show a consistent improvement in both EM and F1 scores across both LongBench(S) and LongBench(M) as the context length increases. This indicates that \textsc{LongFaith} effectively leverages additional contextual information, enhancing the model’s comprehension and reasoning capabilities. These findings validate the scalability of the \textsc{LongFaith} framework in handling long-context scenarios, highlighting its potential for applications that require deep reasoning over extensive inputs.

%% file: conclusion.tex
\section{Conclusion}
\label{sec:conclusion}

This paper addresses the challenge of questionable faithfulness in data synthesis approaches for long-context LLMs.
We propose \textsc{LongFaith}, a novel pipeline synthesizing faithful long-context reasoning datasets through ground truth integration and citation-based reasoning prompts.
Experiments demonstrate its effectiveness, with ablation studies confirming the adaptability of the \textsc{LongFaith}-SFT and \textsc{LongFaith}-PO datasets across diverse long-context scenarios.

%% file: app.tex
\section{An Example of Synthesized \textsc{LongFaith}-SFT and \textsc{LongFaith}-PO Datasets}

\begin{figure}[ht]
    \centering
    \centerline{\includegraphics[width=1.0\linewidth]{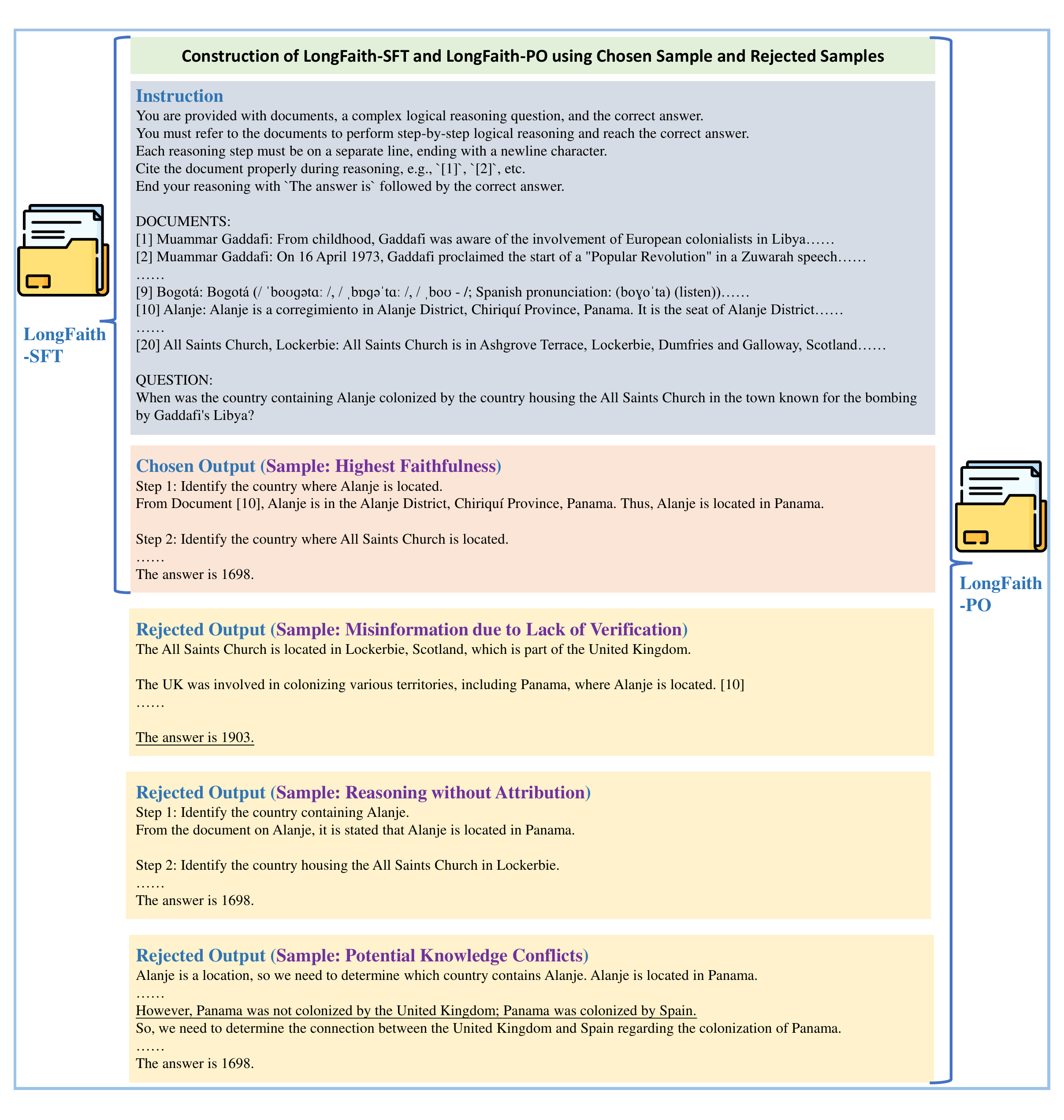}}
    \vspace{-5pt}
    \caption{An Example of synthesized \textsc{LongFaith}-SFT and \textsc{LongFaith}-PO datasets.}
    \label{fig:overview}
    \vspace{-10pt}
\end{figure}

\clearpage

\section{Prompt Templates}
\label{sec:prompt}

\begin{figure}[ht]
    \centering
    \centerline{\includegraphics[width=1.0\linewidth]{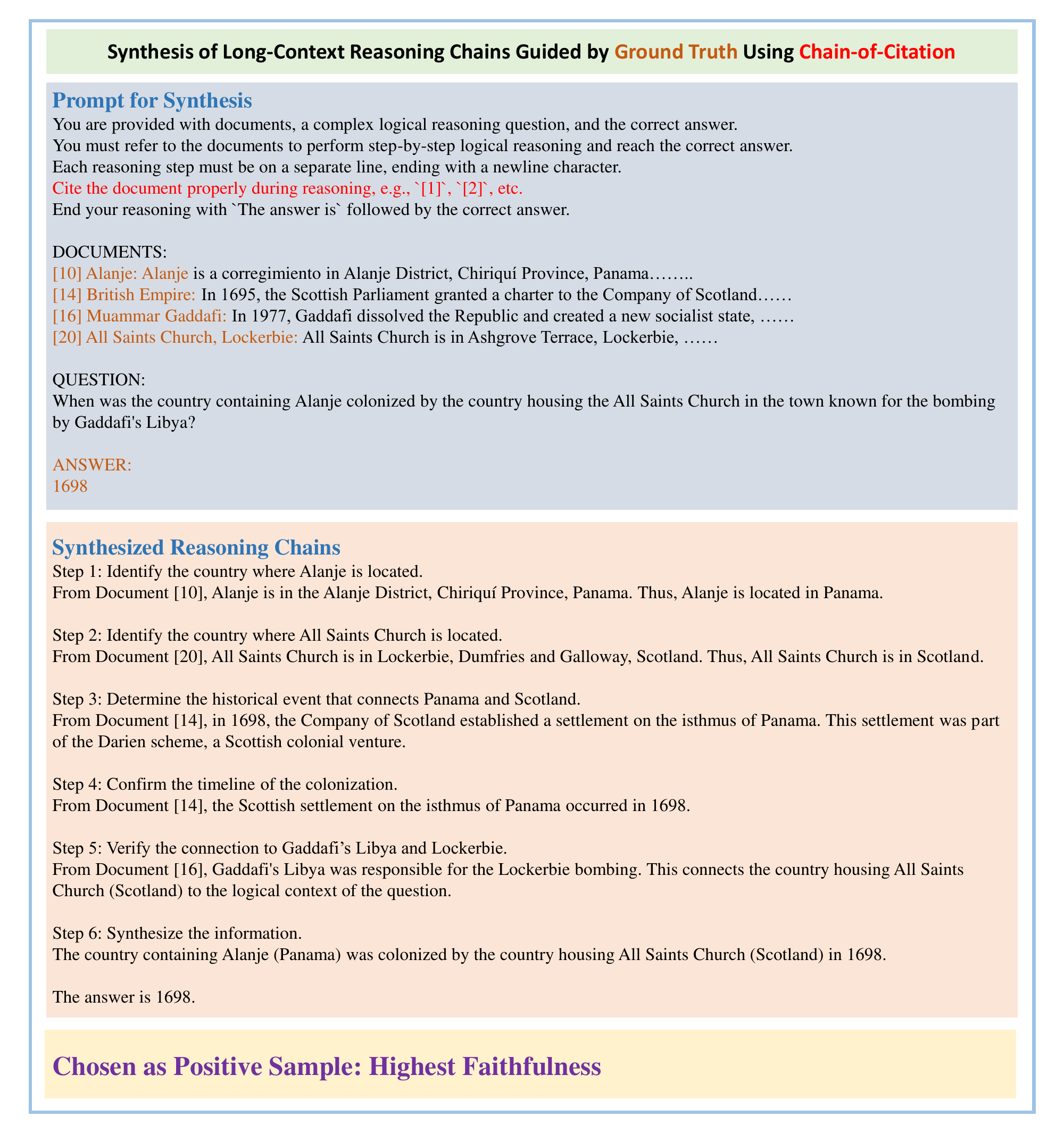}}
    \vspace{-5pt}
    \caption{An example of synthesized chosen reasoning chain. The current reasoning chain must arrive at a correct answer, and reasoning with proper citation proposes more faithfulness and interpretability. Therefore, \textsc{LongFaith} will choose it in supervised fine-tuning and preference optimization as positive sample.}
    \label{fig:pt_1}
    \vspace{-10pt}
\end{figure}

\begin{figure}[ht]
    \centering
    \centerline{\includegraphics[width=1.0\linewidth]{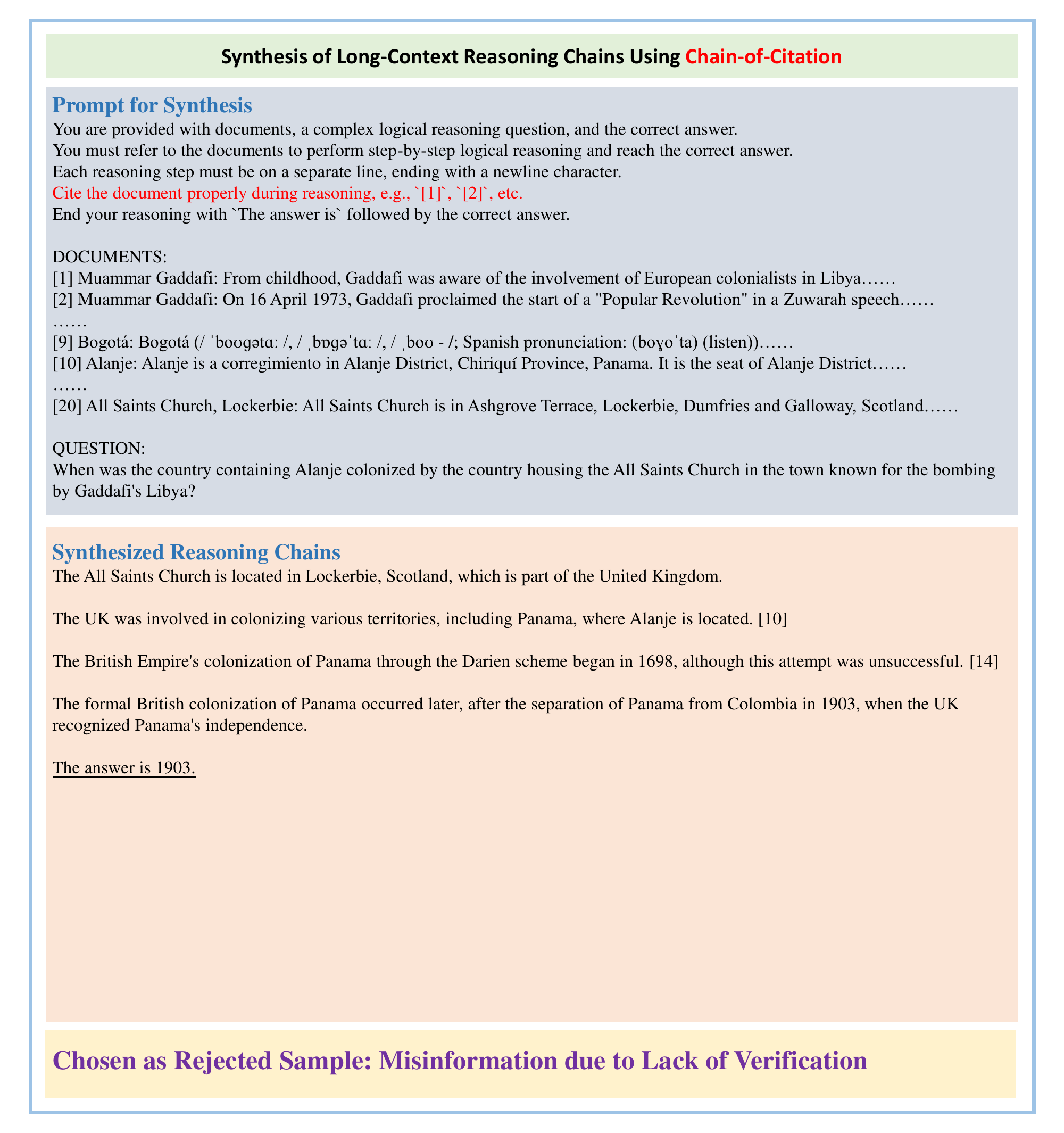}}
    \vspace{-5pt}
    \caption{An example of synthesized rejected reasoning chain. Misinformation due to lack of verification will cause more hallucination if we use current reasoning chain to fine-tune a LLM. Therefore, \textsc{LongFaith} will reject it in preference optimization.}
    \label{fig:pt_2}
    \vspace{-10pt}
\end{figure}

\begin{figure}[ht]
    \centering
    \centerline{\includegraphics[width=1.0\linewidth]{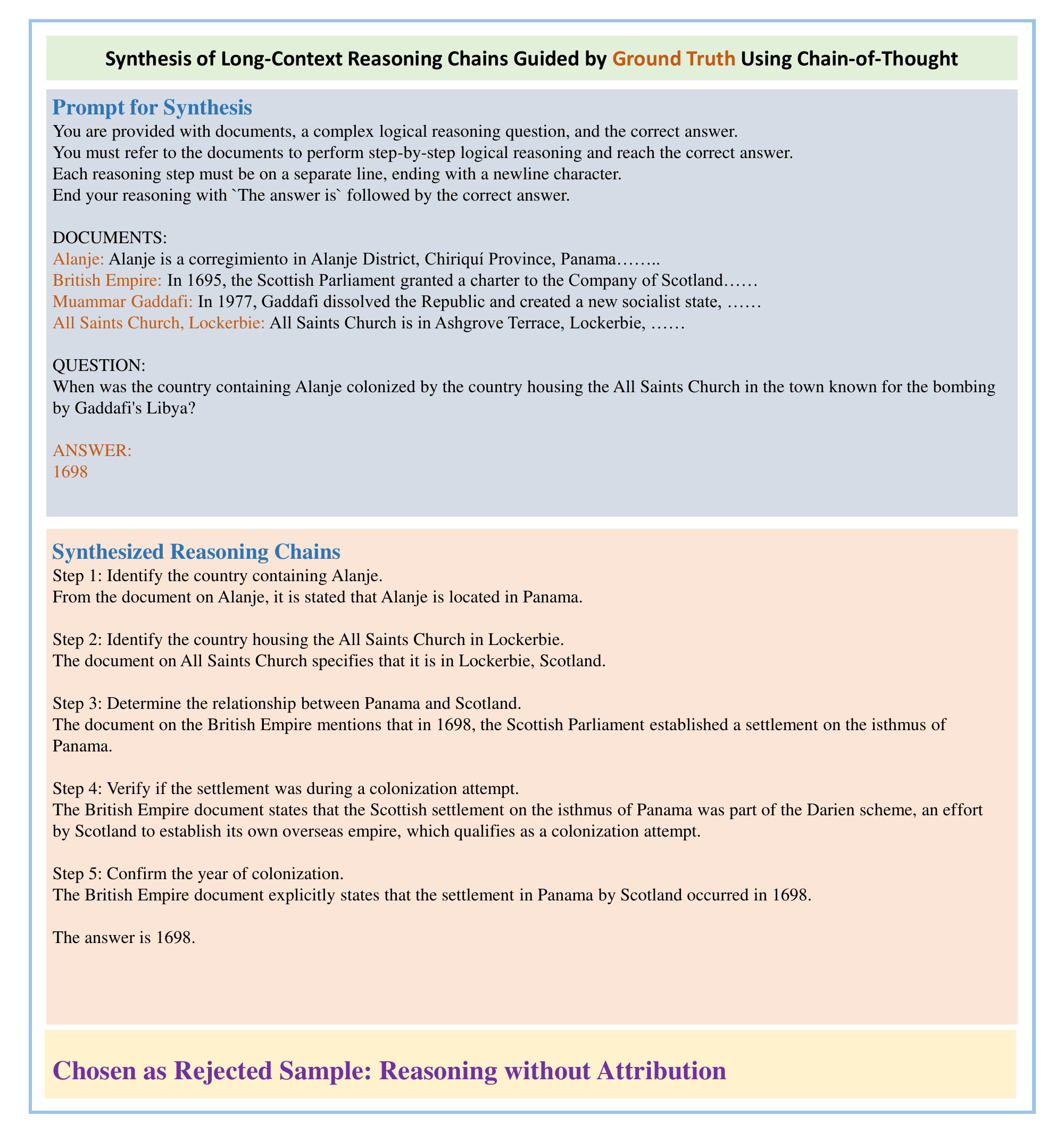}}
    \vspace{-5pt}
    \caption{An example of synthesized rejected reasoning chain. As it mentioned in previous work, lack of attribution will lead to much more interpretability and faithfulness, and response with citation is encouraged. Therefore, \textsc{LongFaith} will reject it in preference optimization.}
    \label{fig:pt_3}
    \vspace{-10pt}
\end{figure}

\begin{figure}[ht]
    \centering
    \centerline{\includegraphics[width=1.0\linewidth]{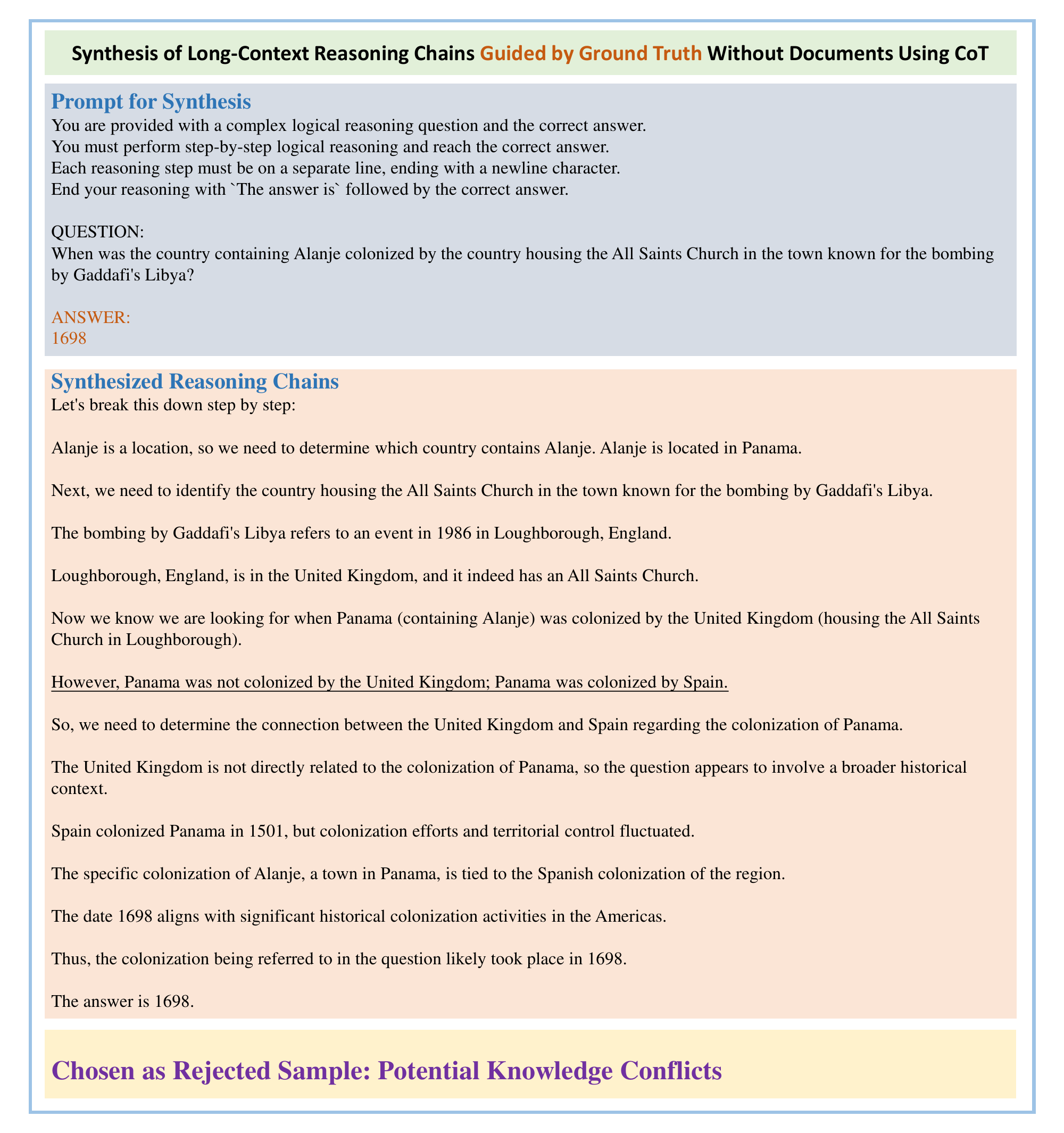}}
    \vspace{-5pt}
    \caption{An example of synthesized rejected reasoning chain. Information related to "Panama was colonized by Spain" is not mentioned in documents provided, which will cause knowledge conflicts if using current reasoning chain for fine-tuning. Therefore, \textsc{LongFaith} will reject it in preference optimization.}
    \label{fig:pt_4}
    \vspace{-10pt}
\end{figure}

We present the prompt templates that are used to synthesize the datasets. The core prompt template that generates long-context reasoning chains guided by ground truth using chain-of-citation is shown in Fig.~\ref{fig:pt_1}. The samples are used in \textsc{LongFaith}-SFT dataset and are chosen as positive in \textsc{LongFaith}-PO dataset, since they are of the highest faithfulness. The other three prompt templates synthesize rejected samples for \textsc{LongFaith}-PO dataset, corresponding to (1) Misinformation due to lack of verification ( Fig.~\ref{fig:pt_2}), (2) Reasoning without attribution (Fig.~\ref{fig:pt_3}), and (3) Potential knowledge conflicts (Fig.~\ref{fig:pt_4}).

\clearpage

\section{Case Study of Metric "Hack" on SubEM comparing previous study with \textsc{LongFaith}}

\begin{figure}[ht]
    \centering
    \centerline{\includegraphics[width=0.83\linewidth]{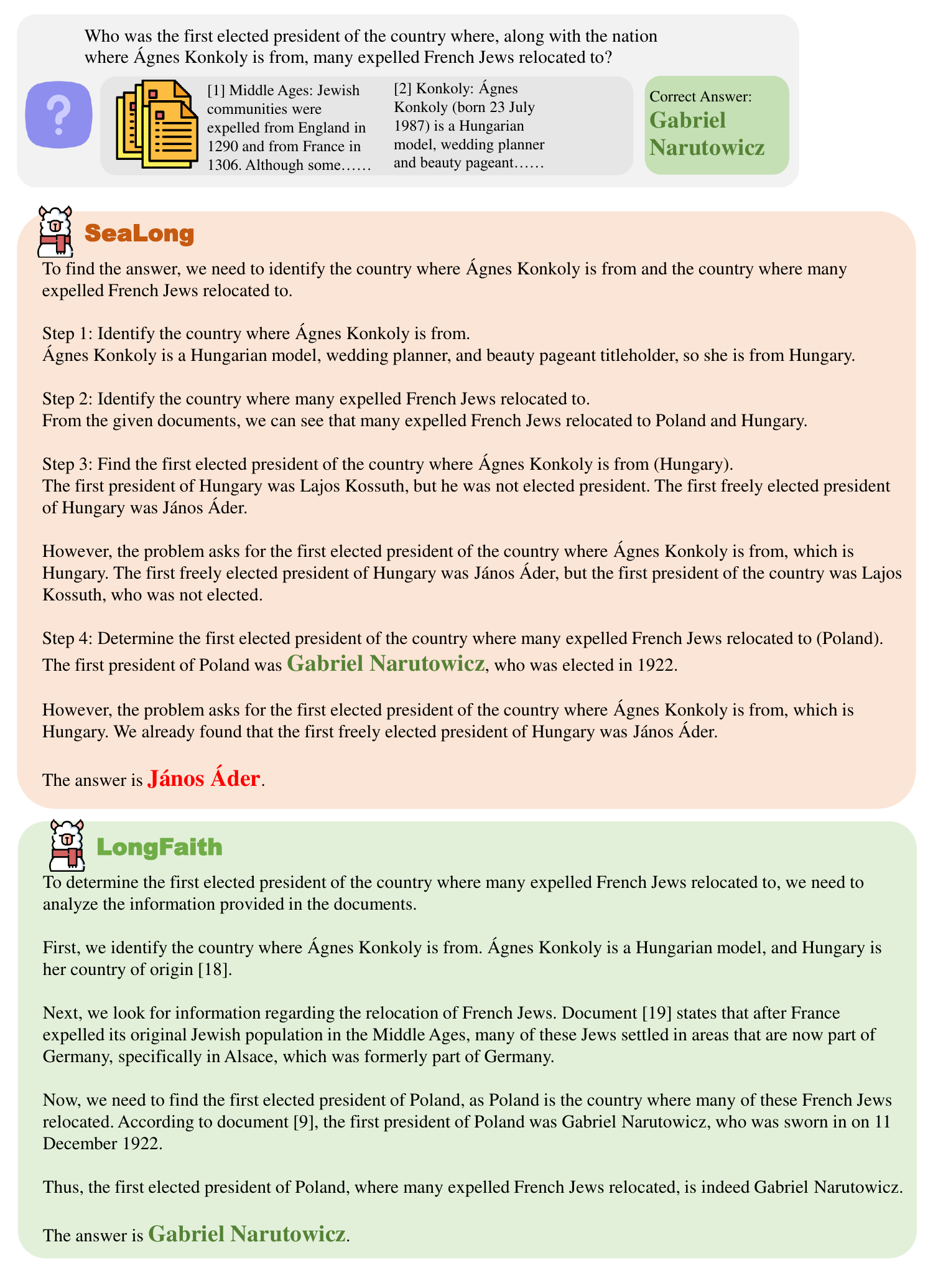}}
    \vspace{-5pt}
    \caption{A case study that SubEM metric is "hacked" by previous study, which conduct more exploration with redundancy in response. \textsc{LongFaith} can arrive at the final correct answer with shorter response.}
    \label{fig:hack}
    \vspace{-10pt}
\end{figure}

\clearpage

\section{Post-Training}
\label{sec:method:post}

In this section, we present two post-training algorithms—Supervised Fine-Tuning (SFT) and Preference Optimization (PO)—to better leverage synthetic data for efficiently enhancing model performance. Specifically, the model performs supervised fine-tuning on high-quality faithful outputs or is trained through reinforcement learning using synthetic preference pairs.

\paragraph{Supervised Fine-tuning on Faithful Outputs}
We minimize the negative log-likelihood of the output as follows:
\begin{equation}
\begin{aligned}    
    \mathcal{L}_{\text{SFT}}
    &= -\frac{1}{|y|} \log \pi_{\theta}(y \;\vert\; x) \\
    &= -\frac{1}{|y|} \sum_{i=1}^{|y|} \log \pi_{\theta}(y_i \;\vert\; x,y_{<i})
\end{aligned}
\end{equation}
where $y$ denotes the high-quality faithful outputs, which are synthesized in Section~\ref{sec:method}.

\paragraph{Reinforcement Learning from Synthetic Preference}

Additionally, we can leverage synthetic preference pairs for reinforcement learning (RL) to fine-tune the model toward generating faithful outputs while reducing the likelihood of low-scoring outputs. Standard RL algorithms for optimizing LLMs include Proximal Policy Optimization (PPO)~\cite{ppo}, RLOO~\cite{RLOO}. However, these methods incur high computational costs. Recent approaches such as Direct Preference Optimization (DPO)~\cite{dpo}, Kahneman-Tversky Optimization (KTO)~\cite{kto}, and Odds Ratio Preference Optimization (ORPO)~\cite{orpo} have been proposed to mitigate both computational and data requirements. In this work, we adopt the ORPO algorithm, which achieves an optimal balance between computational efficiency and model performance.

ORPO introduces an odds ratio loss $\mathcal{L}_{OR}$ that minimizes the negative log odds ratio between preferred ("win" $y_{w}$) and dispreferred ("lose" $y_{l}$) outputs:
\begin{equation}
    \mathcal{L}_{\text{OR}}=-\log \sigma \left( \log \frac{\text{odds}_{\theta}(y_{w}\vert x)}{\text{odds}_{\theta}(y_{l}\vert x)} \right)
\end{equation}
where \(\sigma\) denotes the sigmoid function and \(\text{odds}_{\theta}(y\vert x) = \frac{\pi_{\theta}(y\vert x)}{1 - \pi_{\theta}(y\vert x)}\) measures how much more likely \(y\) is to be generated. The final objective of ORPO is to combine the SFT loss and the OR loss while controlling their relative importance through a hyperparameter \(\beta\):

\begin{equation}
    \mathcal{L}_{\text{ORPO}}=\mathcal{L}_{\text{SFT}} + \beta \cdot \mathcal{L}_{\text{OR}}
\end{equation}

In this paper, the chosen output \(y_w\) is synthesized by LongFaith through comprehensive consideration of Supporting Docs, Chain-of-Citation (CoC), and Ground Truth (GT), and is consequently assigned a high score. Conversely, the rejected output \(y_l\) refers to synthesized outputs that lack at least one of these three critical elements (Supporting Docs, CoC, or GT), which are deemed low-scoring due to insufficient design considerations.

\section{Statistics of Main Experiments}

\begin{table}[ht]
    \centering
    \small
    \begin{tabular}{lcccc}
        \toprule
        \multicolumn{1}{c}{\textbf{\musique}} & \textbf{\#2-Hop} & \textbf{\#3-Hop} & \textbf{\#4-Hop} \\
        \midrule
        \textit{1K}    & 0 & 512 & 512 \\
        \textit{2K}    & 512 & 512 & 1024 \\
        \textit{4K}    & 1024 & 2048 & 1024 \\
        \textit{8K}    & 3072 & 4096 & 1024 \\
        \bottomrule
    \end{tabular}
    \caption{Statistics of train set for synthesis in different size sampled from \textbf{\musique}~\cite{musique}. }
    \label{tab:train_stat}
\end{table}

\begin{table}[ht]
    \centering
    \small
    \begin{tabular}{lccc}
        \toprule
        \multicolumn{1}{c}{\textbf{Datasets}} & \textbf{\#Count} & \textbf{Avg. L.} & \textbf{Max L.} \\
        \midrule
        \multicolumn{4}{c}{\textit{Multi-Hop Reasoning}} \\
        \midrule
        \musique & 500 & & \\
        2-Hop    & 254 & 10843.3 & 17560 \\
        3-Hop    & 154 & 11456.5 & 19225 \\
        4-Hop    & 92  & 11224.3 & 16756 \\
        \midrule
        \twowiki & 500 & & \\
        2-Hop    & 395 & 4449.5 & 10631 \\
        4-Hop    & 105 & 4041.4 & 9365  \\
        \midrule
        \hotpot    & 500 & & \\
        Bridge     & 412 & 6301.0 & 15702 \\
        Comparison & 88  & 5777.6 & 11939 \\
        \midrule
        \multicolumn{4}{c}{\textit{LongBench}} \\
        \midrule
        Qasper (S)          & 200 & 24262.3 & 101636 \\
        MultiFieldQA-En (S) & 150 & 29583.7 & 64751 \\
        MuSiQue (M)         & 200 & 69876.8 & 82338 \\
        2WikiMHQA (M)           & 200 & 30076.5 & 72971 \\
        HotpotQA (M)        & 200 & 57041.4 & 81815 \\
        \bottomrule
    \end{tabular}
    \caption{Statistics of test sets including three long-context multi-hop reasoning datasets sampled by~\cite{multihopdatasets} and five long-context QA datasets from LongBench~\cite{longbench}. \textbf{Avg. L.} and \textbf{Max L. } refer to the average length and max length of input prompts for test samples. \textbf{(S)} and \textbf{(M)} refer to Single-doc QA and Multi-doc QA in LongBench.}
    \label{tab:dataset_stat}
\end{table}

\begin{table}[ht]
    \centering
    \small
    \begin{tabular}{ccccc}
        \toprule
        \multicolumn{1}{c}{\textbf{Datasets}} & \textbf{Instruction} & \textbf{Output(Chosen)} & \textbf{Rejected} \\
        \midrule
        LongAlpaca    & 52043.2 & 620.7 & 0 \\
        LongAlign    & 36307.2 & 1412.6 & 0 \\
        \musique-Attribute & 11395.0 & 343.7 & 0 \\
        LongMIT    & 280808.9 & 825.2 & 0 \\
        LongReward & 72892.2 & 913.4 & 960.6 \\
        \textsc{SeaLong} & 82248.6 & 1156.5 & 1139.1 \\
        \textsc{LongFaith} & 11542.1 & 1029.6 & 896.7 \\
        \bottomrule
    \end{tabular}
    \caption{Average text length of baseline datasets and \textsc{LongFaith} in main experiments in Tab.~\ref{tab:longbench_exp}. All of them has \textit{2K} examples.}
    \label{tab:baseline_stat}
\end{table}

\begin{table}[ht]
    \centering
    \small
    \begin{tabular}{ccccccc}
        \toprule
        \textbf{Datasets} & \textbf{\musique} & \textbf{\twowiki} & \textbf{HotpotQA} & \textbf{Qasper} & \textbf{MultiFieldQA} & \textbf{Avg.L} \\
        \midrule
        LongAlpaca      & 365.62  & 372.97   & 319.60   & 657.34  & 511.25  & 445.36 \\
        LongAlign       & 493.56  & 349.65   & 371.77   & 651.76  & 623.15  & 497.98 \\
        \musique-Attribute & 99.61   & 168.74   & 164.24   & 317.75  & 252.95  & 200.66 \\
        LongMIT         & 138.03  & 159.16   & 116.40   & 194.41  & 196.43  & 160.89 \\
        LongReward-SFT  & 285.20  & 241.47   & 178.26   & 750.83  & 537.95  & 398.74 \\
        SeaLong-SFT     & 1091.54 & 776.01   & 926.29   & 1035.77 & 822.82  & 930.49 \\
        LongFaith-SFT   & 820.04  & 619.18   & 771.68   & 1056.55 & 941.13  & 841.72 \\
        LongReward-PO   & 219.90  & 253.41   & 179.44   & 616.11  & 460.40  & 345.85 \\
        SeaLong-PO      & 961.51  & 740.14   & 891.75   & 946.68  & 826.77  & 873.37 \\
        LongFaith-PO    & 831.17  & 669.76   & 786.77   & 1034.71 & 917.11  & 847.90 \\
        \bottomrule
    \end{tabular}
    \caption{Average length of model output in test sets trained on different synthesized instruction.}
    \label{tab:output_stat}
\end{table}

\clearpage

\section{Hyperparameters}
\label{sec:hyperparameters}

\begin{table}[ht]
\centering
\begin{tabular}{c c}
\toprule
\textbf{Hyperparameters}        & \textbf{Value} \\
\midrule
\# GPUs used                    & 4                \\
Learning rate                   & 5e-5             \\
Per-device batch size           & 1                \\
Gradient accumulation steps     & 8                \\
LoRA rank                       & 32               \\
LoRA alpha                      & 64               \\
LoRA dropout                    & 0.1              \\
ORPO beta                       & 0.1              \\
Warm-up ratio                   & 0.1              \\
Epochs                          & 1                \\
Precision                       & bfloat16         \\
Optimizer                       & AdamW            \\
\bottomrule
\end{tabular}
\caption{Hyperparameter settings of fine-tuning and preference optimization.}
\label{tab:hyperparameters}
\end{table}


%% file: acl_latex.bbl
\begin{thebibliography}{60}
\providecommand{\natexlab}[1]{#1}

\bibitem[{Ahmadian et~al.(2024)Ahmadian, Cremer, Gallé, Fadaee, Kreutzer, Pietquin, Üstün, and Hooker}]{RLOO}
Arash Ahmadian, Chris Cremer, Matthias Gallé, Marzieh Fadaee, Julia Kreutzer, Olivier Pietquin, Ahmet Üstün, and Sara Hooker. 2024.
\newblock Back to basics: Revisiting reinforce style optimization for learning from human feedback in llms.
\newblock \emph{arXiv preprint arXiv: 2402.14740}.

\bibitem[{An et~al.(2024)An, Ma, Lin, Zheng, and Lou}]{in2}
Shengnan An, Zexiong Ma, Zeqi Lin, Nanning Zheng, and Jian-Guang Lou. 2024.
\newblock Make your llm fully utilize the context.
\newblock \emph{arXiv preprint arXiv:2404.16811}.

\bibitem[{Bai et~al.(2024{\natexlab{a}})Bai, Lv, Zhang, He, Qi, Hou, Tang, Dong, and Li}]{longalign}
Yushi Bai, Xin Lv, Jiajie Zhang, Yuze He, Ji~Qi, Lei Hou, Jie Tang, Yuxiao Dong, and Juanzi Li. 2024{\natexlab{a}}.
\newblock Longalign: A recipe for long context alignment of large language models.
\newblock \emph{arXiv preprint arXiv:2401.18058}.

\bibitem[{Bai et~al.(2023)Bai, Lv, Zhang, Lyu, Tang, Huang, Du, Liu, Zeng, Hou et~al.}]{longbench}
Yushi Bai, Xin Lv, Jiajie Zhang, Hongchang Lyu, Jiankai Tang, Zhidian Huang, Zhengxiao Du, Xiao Liu, Aohan Zeng, Lei Hou, et~al. 2023.
\newblock Longbench: A bilingual, multitask benchmark for long context understanding.
\newblock \emph{arXiv preprint arXiv:2308.14508}.

\bibitem[{Bai et~al.(2024{\natexlab{b}})Bai, Tu, Zhang, Peng, Wang, Lv, Cao, Xu, Hou, Dong et~al.}]{longbenchv2}
Yushi Bai, Shangqing Tu, Jiajie Zhang, Hao Peng, Xiaozhi Wang, Xin Lv, Shulin Cao, Jiazheng Xu, Lei Hou, Yuxiao Dong, et~al. 2024{\natexlab{b}}.
\newblock Longbench v2: Towards deeper understanding and reasoning on realistic long-context multitasks.
\newblock \emph{arXiv preprint arXiv:2412.15204}.

\bibitem[{Beltagy et~al.(2020)Beltagy, Peters, and Cohan}]{longformer}
Iz~Beltagy, Matthew~E Peters, and Arman Cohan. 2020.
\newblock Longformer: The long-document transformer.
\newblock \emph{arXiv preprint arXiv:2004.05150}.

\bibitem[{Berchansky et~al.(2024)Berchansky, Fleischer, Wasserblat, and Izsak}]{cotar}
Moshe Berchansky, Daniel Fleischer, Moshe Wasserblat, and Peter Izsak. 2024.
\newblock Cotar: Chain-of-thought attribution reasoning with multi-level granularity.
\newblock \emph{arXiv preprint arXiv:2404.10513}.

\bibitem[{Chen et~al.(2023{\natexlab{a}})Chen, Wong, Chen, and Tian}]{extending}
Shouyuan Chen, Sherman Wong, Liangjian Chen, and Yuandong Tian. 2023{\natexlab{a}}.
\newblock Extending context window of large language models via positional interpolation.
\newblock \emph{arXiv preprint arXiv:2306.15595}.

\bibitem[{Chen et~al.(2023{\natexlab{b}})Chen, Qian, Tang, Lai, Liu, Han, and Jia}]{longlora}
Yukang Chen, Shengju Qian, Haotian Tang, Xin Lai, Zhijian Liu, Song Han, and Jiaya Jia. 2023{\natexlab{b}}.
\newblock Longlora: Efficient fine-tuning of long-context large language models.
\newblock \emph{arXiv preprint arXiv:2309.12307}.

\bibitem[{Chen et~al.(2024)Chen, Chen, Qin, Guo, Lv, Zou, Che, Yan, Chen, and Lin}]{longmit}
Zhi Chen, Qiguang Chen, Libo Qin, Qipeng Guo, Haijun Lv, Yicheng Zou, Wanxiang Che, Hang Yan, Kai Chen, and Dahua Lin. 2024.
\newblock What are the essential factors in crafting effective long context multi-hop instruction datasets? insights and best practices.
\newblock \emph{arXiv preprint arXiv:2409.01893}.

\bibitem[{Choi et~al.(2018)Choi, He, Iyyer, Yatskar, Yih, Choi, Liang, and Zettlemoyer}]{quac}
Eunsol Choi, He~He, Mohit Iyyer, Mark Yatskar, Wen-tau Yih, Yejin Choi, Percy Liang, and Luke Zettlemoyer. 2018.
\newblock Quac: Question answering in context.
\newblock \emph{arXiv preprint arXiv:1808.07036}.

\bibitem[{Dasigi et~al.(2021)Dasigi, Lo, Beltagy, Cohan, Smith, and Gardner}]{qasper}
Pradeep Dasigi, Kyle Lo, Iz~Beltagy, Arman Cohan, Noah~A Smith, and Matt Gardner. 2021.
\newblock A dataset of information-seeking questions and answers anchored in research papers.
\newblock \emph{arXiv preprint arXiv:2105.03011}.

\bibitem[{Ding et~al.(2023)Ding, Ma, Dong, Zhang, Huang, Wang, Zheng, and Wei}]{longnet}
Jiayu Ding, Shuming Ma, Li~Dong, Xingxing Zhang, Shaohan Huang, Wenhui Wang, Nanning Zheng, and Furu Wei. 2023.
\newblock Longnet: Scaling transformers to 1,000,000,000 tokens.
\newblock \emph{arXiv preprint arXiv:2307.02486}.

\bibitem[{Ding et~al.(2024)Ding, Zhang, Zhang, Xu, Shang, Xu, Yang, and Yang}]{longrope}
Yiran Ding, Li~Lyna Zhang, Chengruidong Zhang, Yuanyuan Xu, Ning Shang, Jiahang Xu, Fan Yang, and Mao Yang. 2024.
\newblock Longrope: Extending llm context window beyond 2 million tokens.
\newblock \emph{arXiv preprint arXiv:2402.13753}.

\bibitem[{Dubey et~al.(2024)Dubey, Jauhri, Pandey, Kadian, Al-Dahle, Letman, Mathur, Schelten, Yang, Fan et~al.}]{llama3}
Abhimanyu Dubey, Abhinav Jauhri, Abhinav Pandey, Abhishek Kadian, Ahmad Al-Dahle, Aiesha Letman, Akhil Mathur, Alan Schelten, Amy Yang, Angela Fan, et~al. 2024.
\newblock The llama 3 herd of models.
\newblock \emph{arXiv preprint arXiv:2407.21783}.

\bibitem[{Ethayarajh et~al.(2024)Ethayarajh, Xu, Muennighoff, Jurafsky, and Kiela}]{kto}
Kawin Ethayarajh, Winnie Xu, Niklas Muennighoff, Dan Jurafsky, and Douwe Kiela. 2024.
\newblock Kto: Model alignment as prospect theoretic optimization.
\newblock \emph{International Conference on Machine Learning}.

\bibitem[{Fierro et~al.(2024)Fierro, Amplayo, Huot, De~Cao, Maynez, Narayan, and Lapata}]{learning}
Constanza Fierro, Reinald~Kim Amplayo, Fantine Huot, Nicola De~Cao, Joshua Maynez, Shashi Narayan, and Mirella Lapata. 2024.
\newblock Learning to plan and generate text with citations.
\newblock \emph{arXiv preprint arXiv:2404.03381}.

\bibitem[{Fu et~al.(2024)Fu, Panda, Niu, Yue, Hajishirzi, Kim, and Peng}]{dataengineering}
Yao Fu, Rameswar Panda, Xinyao Niu, Xiang Yue, Hannaneh Hajishirzi, Yoon Kim, and Hao Peng. 2024.
\newblock Data engineering for scaling language models to 128k context.
\newblock \emph{arXiv preprint arXiv:2402.10171}.

\bibitem[{Gao et~al.(2024)Gao, Wettig, Yen, and Chen}]{prolong}
Tianyu Gao, Alexander Wettig, Howard Yen, and Danqi Chen. 2024.
\newblock How to train long-context language models (effectively).
\newblock \emph{arXiv preprint arXiv:2410.02660}.

\bibitem[{Gao et~al.(2023)Gao, Yen, Yu, and Chen}]{alce}
Tianyu Gao, Howard Yen, Jiatong Yu, and Danqi Chen. 2023.
\newblock Enabling large language models to generate text with citations.
\newblock \emph{arXiv preprint arXiv:2305.14627}.

\bibitem[{Gu et~al.(2024)Gu, Jiang, Shi, Tan, Zhai, Xu, Li, Shen, Ma, Liu et~al.}]{llmasajudge}
Jiawei Gu, Xuhui Jiang, Zhichao Shi, Hexiang Tan, Xuehao Zhai, Chengjin Xu, Wei Li, Yinghan Shen, Shengjie Ma, Honghao Liu, et~al. 2024.
\newblock A survey on llm-as-a-judge.
\newblock \emph{arXiv preprint arXiv:2411.15594}.

\bibitem[{Han et~al.(2024)Han, Wang, Peng, Xiong, Chen, Ji, and Wang}]{lminfinite}
Chi Han, Qifan Wang, Hao Peng, Wenhan Xiong, Yu~Chen, Heng Ji, and Sinong Wang. 2024.
\newblock Lm-infinite: Zero-shot extreme length generalization for large language models.
\newblock In \emph{Proceedings of the 2024 Conference of the North American Chapter of the Association for Computational Linguistics: Human Language Technologies (Volume 1: Long Papers)}, pages 3991--4008.

\bibitem[{Ho et~al.(2020)Ho, Nguyen, Sugawara, and Aizawa}]{twowiki}
Xanh Ho, Anh-Khoa~Duong Nguyen, Saku Sugawara, and Akiko Aizawa. 2020.
\newblock Constructing a multi-hop qa dataset for comprehensive evaluation of reasoning steps.
\newblock \emph{arXiv preprint arXiv:2011.01060}.

\bibitem[{Hong et~al.(2024)Hong, Lee, and Thorne}]{orpo}
Jiwoo Hong, Noah Lee, and James Thorne. 2024.
\newblock Orpo: Monolithic preference optimization without reference model.
\newblock In \emph{Proceedings of the 2024 Conference on Empirical Methods in Natural Language Processing}, pages 11170--11189.

\bibitem[{Hsieh et~al.(2024)Hsieh, Sun, Kriman, Acharya, Rekesh, Jia, Zhang, and Ginsburg}]{ruler}
Cheng-Ping Hsieh, Simeng Sun, Samuel Kriman, Shantanu Acharya, Dima Rekesh, Fei Jia, Yang Zhang, and Boris Ginsburg. 2024.
\newblock Ruler: What's the real context size of your long-context language models?
\newblock \emph{arXiv preprint arXiv:2404.06654}.

\bibitem[{Hu et~al.(2021)Hu, Shen, Wallis, Allen-Zhu, Li, Wang, Wang, and Chen}]{lora}
Edward~J Hu, Yelong Shen, Phillip Wallis, Zeyuan Allen-Zhu, Yuanzhi Li, Shean Wang, Lu~Wang, and Weizhu Chen. 2021.
\newblock Lora: Low-rank adaptation of large language models.
\newblock \emph{arXiv preprint arXiv:2106.09685}.

\bibitem[{Huang et~al.(2024)Huang, Feng, Ma, Gu, Zhong, Feng, Yu, Peng, Tang, Tu et~al.}]{finegrained}
Lei Huang, Xiaocheng Feng, Weitao Ma, Yuxuan Gu, Weihong Zhong, Xiachong Feng, Weijiang Yu, Weihua Peng, Duyu Tang, Dandan Tu, et~al. 2024.
\newblock Learning fine-grained grounded citations for attributed large language models.
\newblock \emph{arXiv preprint arXiv:2408.04568}.

\bibitem[{Huang et~al.(2023)Huang, Yu, Ma, Zhong, Feng, Wang, Chen, Peng, Feng, Qin et~al.}]{hallu}
Lei Huang, Weijiang Yu, Weitao Ma, Weihong Zhong, Zhangyin Feng, Haotian Wang, Qianglong Chen, Weihua Peng, Xiaocheng Feng, Bing Qin, et~al. 2023.
\newblock A survey on hallucination in large language models: Principles, taxonomy, challenges, and open questions.
\newblock \emph{arXiv preprint arXiv:2311.05232}.

\bibitem[{Hurst et~al.(2024)Hurst, Lerer, Goucher, Perelman, Ramesh, Clark, Ostrow, Welihinda, Hayes, Radford et~al.}]{gpt-4o}
Aaron Hurst, Adam Lerer, Adam~P Goucher, Adam Perelman, Aditya Ramesh, Aidan Clark, AJ~Ostrow, Akila Welihinda, Alan Hayes, Alec Radford, et~al. 2024.
\newblock Gpt-4o system card.
\newblock \emph{arXiv preprint arXiv:2410.21276}.

\bibitem[{Jiang et~al.(2025)Jiang, Ma, Xu, Yang, Zhang, and Guo}]{sog}
Xuhui Jiang, Shengjie Ma, Chengjin Xu, Cehao Yang, Liyu Zhang, and Jian Guo. 2025.
\newblock Synthesize-on-graph: Knowledgeable synthetic data generation for continue pre-training of large language models.
\newblock \emph{arXiv preprint arXiv:2505.00979}.

\bibitem[{Kwon et~al.(2023)Kwon, Li, Zhuang, Sheng, Zheng, Yu, Gonzalez, Zhang, and Stoica}]{vllm}
Woosuk Kwon, Zhuohan Li, Siyuan Zhuang, Ying Sheng, Lianmin Zheng, Cody~Hao Yu, Joseph~E. Gonzalez, Hao Zhang, and Ion Stoica. 2023.
\newblock Efficient memory management for large language model serving with pagedattention.
\newblock In \emph{Proceedings of the ACM SIGOPS 29th Symposium on Operating Systems Principles}.

\bibitem[{Levy et~al.(2024)Levy, Jacoby, and Goldberg}]{sametaskmoretokens}
Mosh Levy, Alon Jacoby, and Yoav Goldberg. 2024.
\newblock Same task, more tokens: the impact of input length on the reasoning performance of large language models.
\newblock \emph{arXiv preprint arXiv:2402.14848}.

\bibitem[{Li et~al.(2023)Li, Sun, Hu, Liu, Chen, Hu, Wu, and Zhang}]{attribution}
Dongfang Li, Zetian Sun, Xinshuo Hu, Zhenyu Liu, Ziyang Chen, Baotian Hu, Aiguo Wu, and Min Zhang. 2023.
\newblock A survey of large language models attribution.
\newblock \emph{arXiv preprint arXiv:2311.03731}.

\bibitem[{Li et~al.(2024{\natexlab{a}})Li, Yang, Cheng, Liu, Yu, Yang, and Lam}]{sealong}
Siheng Li, Cheng Yang, Zesen Cheng, Lemao Liu, Mo~Yu, Yujiu Yang, and Wai Lam. 2024{\natexlab{a}}.
\newblock Large language models can self-improve in long-context reasoning.
\newblock \emph{arXiv preprint arXiv:2411.08147}.

\bibitem[{Li et~al.(2024{\natexlab{b}})Li, Liang, Lyu, and Wang}]{coc}
Yanyang Li, Shuo Liang, Michael~R Lyu, and Liwei Wang. 2024{\natexlab{b}}.
\newblock Making long-context language models better multi-hop reasoners.
\newblock \emph{arXiv preprint arXiv:2408.03246}.

\bibitem[{Liu et~al.(2024)Liu, Lin, Hewitt, Paranjape, Bevilacqua, Petroni, and Liang}]{lostinthemiddle}
Nelson~F Liu, Kevin Lin, John Hewitt, Ashwin Paranjape, Michele Bevilacqua, Fabio Petroni, and Percy Liang. 2024.
\newblock Lost in the middle: How language models use long contexts.
\newblock \emph{Transactions of the Association for Computational Linguistics}, 12:157--173.

\bibitem[{Ma et~al.(2024)Ma, Xu, Jiang, Li, Qu, Yang, Mao, and Guo}]{tog2}
Shengjie Ma, Chengjin Xu, Xuhui Jiang, Muzhi Li, Huaren Qu, Cehao Yang, Jiaxin Mao, and Jian Guo. 2024.
\newblock Think-on-graph 2.0: Deep and faithful large language model reasoning with knowledge-guided retrieval augmented generation.
\newblock \emph{arXiv preprint arXiv:2407.10805}.

\bibitem[{Peng et~al.(2023)Peng, Quesnelle, Fan, and Shippole}]{yarn}
Bowen Peng, Jeffrey Quesnelle, Honglu Fan, and Enrico Shippole. 2023.
\newblock Yarn: Efficient context window extension of large language models.
\newblock \emph{arXiv preprint arXiv:2309.00071}.

\bibitem[{Rafailov et~al.(2024)Rafailov, Sharma, Mitchell, Manning, Ermon, and Finn}]{dpo}
Rafael Rafailov, Archit Sharma, Eric Mitchell, Christopher~D Manning, Stefano Ermon, and Chelsea Finn. 2024.
\newblock Direct preference optimization: Your language model is secretly a reward model.
\newblock \emph{Advances in Neural Information Processing Systems}, 36.

\bibitem[{Schulman et~al.(2017)Schulman, Wolski, Dhariwal, Radford, and Klimov}]{ppo}
John Schulman, Filip Wolski, Prafulla Dhariwal, Alec Radford, and Oleg Klimov. 2017.
\newblock Proximal policy optimization algorithms.
\newblock \emph{arXiv preprint arXiv: 1707.06347}.

\bibitem[{Shi et~al.(2023)Shi, Chen, Misra, Scales, Dohan, Chi, Sch{\"a}rli, and Zhou}]{distracted}
Freda Shi, Xinyun Chen, Kanishka Misra, Nathan Scales, David Dohan, Ed~H Chi, Nathanael Sch{\"a}rli, and Denny Zhou. 2023.
\newblock Large language models can be easily distracted by irrelevant context.
\newblock In \emph{International Conference on Machine Learning}, pages 31210--31227. PMLR.

\bibitem[{Sun et~al.(2023)Sun, Xu, Tang, Wang, Lin, Gong, Shum, and Guo}]{tog}
Jiashuo Sun, Chengjin Xu, Lumingyuan Tang, Saizhuo Wang, Chen Lin, Yeyun Gong, Heung-Yeung Shum, and Jian Guo. 2023.
\newblock Think-on-graph: Deep and responsible reasoning of large language model with knowledge graph.
\newblock \emph{arXiv preprint arXiv:2307.07697}.

\bibitem[{Trivedi et~al.(2022{\natexlab{a}})Trivedi, Balasubramanian, Khot, and Sabharwal}]{multihopdatasets}
Harsh Trivedi, Niranjan Balasubramanian, Tushar Khot, and Ashish Sabharwal. 2022{\natexlab{a}}.
\newblock Interleaving retrieval with chain-of-thought reasoning for knowledge-intensive multi-step questions.
\newblock \emph{arXiv preprint arXiv:2212.10509}.

\bibitem[{Trivedi et~al.(2022{\natexlab{b}})Trivedi, Balasubramanian, Khot, and Sabharwal}]{musique}
Harsh Trivedi, Niranjan Balasubramanian, Tushar Khot, and Ashish Sabharwal. 2022{\natexlab{b}}.
\newblock Musique: Multihop questions via single-hop question composition.
\newblock \emph{Transactions of the Association for Computational Linguistics}, 10:539--554.

\bibitem[{Wang et~al.(2022)Wang, Kordi, Mishra, Liu, Smith, Khashabi, and Hajishirzi}]{selfinstruct}
Yizhong Wang, Yeganeh Kordi, Swaroop Mishra, Alisa Liu, Noah~A Smith, Daniel Khashabi, and Hannaneh Hajishirzi. 2022.
\newblock Self-instruct: Aligning language models with self-generated instructions.
\newblock \emph{arXiv preprint arXiv:2212.10560}.

\bibitem[{Wei et~al.(2022)Wei, Wang, Schuurmans, Bosma, Xia, Chi, Le, Zhou et~al.}]{cot}
Jason Wei, Xuezhi Wang, Dale Schuurmans, Maarten Bosma, Fei Xia, Ed~Chi, Quoc~V Le, Denny Zhou, et~al. 2022.
\newblock Chain-of-thought prompting elicits reasoning in large language models.
\newblock \emph{Advances in neural information processing systems}, 35:24824--24837.

\bibitem[{Wu et~al.(2024)Wu, Xie, Chen, Zhu, Zhang, and Xiao}]{irrelevant}
Siye Wu, Jian Xie, Jiangjie Chen, Tinghui Zhu, Kai Zhang, and Yanghua Xiao. 2024.
\newblock How easily do irrelevant inputs skew the responses of large language models?
\newblock \emph{arXiv preprint arXiv:2404.03302}.

\bibitem[{Xiong et~al.(2023)Xiong, Liu, Molybog, Zhang, Bhargava, Hou, Martin, Rungta, Sankararaman, Oguz et~al.}]{effective}
Wenhan Xiong, Jingyu Liu, Igor Molybog, Hejia Zhang, Prajjwal Bhargava, Rui Hou, Louis Martin, Rashi Rungta, Karthik~Abinav Sankararaman, Barlas Oguz, et~al. 2023.
\newblock Effective long-context scaling of foundation models.
\newblock \emph{arXiv preprint arXiv:2309.16039}.

\bibitem[{Xu et~al.(2024)Xu, Qi, Guo, Wang, Wang, Zhang, and Xu}]{knowledgeconflicts}
Rongwu Xu, Zehan Qi, Zhijiang Guo, Cunxiang Wang, Hongru Wang, Yue Zhang, and Wei Xu. 2024.
\newblock Knowledge conflicts for llms: A survey.
\newblock \emph{arXiv preprint arXiv:2403.08319}.

\bibitem[{Yang et~al.(2024)Yang, Yang, Zhang, Hui, Zheng, Yu, Li, Liu, Huang, Wei et~al.}]{qwen2.5}
An~Yang, Baosong Yang, Beichen Zhang, Binyuan Hui, Bo~Zheng, Bowen Yu, Chengyuan Li, Dayiheng Liu, Fei Huang, Haoran Wei, et~al. 2024.
\newblock Qwen2. 5 technical report.
\newblock \emph{arXiv preprint arXiv:2412.15115}.

\bibitem[{Yang et~al.(2018)Yang, Qi, Zhang, Bengio, Cohen, Salakhutdinov, and Manning}]{hotpotqa}
Zhilin Yang, Peng Qi, Saizheng Zhang, Yoshua Bengio, William~W Cohen, Ruslan Salakhutdinov, and Christopher~D Manning. 2018.
\newblock Hotpotqa: A dataset for diverse, explainable multi-hop question answering.
\newblock \emph{arXiv preprint arXiv:1809.09600}.

\bibitem[{Yen et~al.(2024)Yen, Gao, Hou, Ding, Fleischer, Izsak, Wasserblat, and Chen}]{helmet}
Howard Yen, Tianyu Gao, Minmin Hou, Ke~Ding, Daniel Fleischer, Peter Izsak, Moshe Wasserblat, and Danqi Chen. 2024.
\newblock Helmet: How to evaluate long-context language models effectively and thoroughly.
\newblock \emph{arXiv preprint arXiv:2410.02694}.

\bibitem[{Yue et~al.(2023)Yue, Wang, Chen, Zhang, Su, and Sun}]{automatic}
Xiang Yue, Boshi Wang, Ziru Chen, Kai Zhang, Yu~Su, and Huan Sun. 2023.
\newblock Automatic evaluation of attribution by large language models.
\newblock \emph{arXiv preprint arXiv:2305.06311}.

\bibitem[{Zhang et~al.(2024{\natexlab{a}})Zhang, Zhang, Zhang, Yong, and Huang}]{retrieval}
Jiahao Zhang, Haiyang Zhang, Dongmei Zhang, Liu Yong, and Shen Huang. 2024{\natexlab{a}}.
\newblock End-to-end beam retrieval for multi-hop question answering.
\newblock In \emph{Proceedings of the 2024 Conference of the North American Chapter of the Association for Computational Linguistics: Human Language Technologies (Volume 1: Long Papers)}, pages 1718--1731.

\bibitem[{Zhang et~al.(2024{\natexlab{b}})Zhang, Bai, Lv, Gu, Liu, Zou, Cao, Hou, Dong, Feng et~al.}]{longcite}
Jiajie Zhang, Yushi Bai, Xin Lv, Wanjun Gu, Danqing Liu, Minhao Zou, Shulin Cao, Lei Hou, Yuxiao Dong, Ling Feng, et~al. 2024{\natexlab{b}}.
\newblock Longcite: Enabling llms to generate fine-grained citations in long-context qa.
\newblock \emph{arXiv preprint arXiv:2409.02897}.

\bibitem[{Zhang et~al.(2024{\natexlab{c}})Zhang, Hou, Lv, Cao, Hou, Niu, Hou, Dong, Feng, and Li}]{longreward}
Jiajie Zhang, Zhongni Hou, Xin Lv, Shulin Cao, Zhenyu Hou, Yilin Niu, Lei Hou, Yuxiao Dong, Ling Feng, and Juanzi Li. 2024{\natexlab{c}}.
\newblock Longreward: Improving long-context large language models with ai feedback.
\newblock \emph{arXiv preprint arXiv:2410.21252}.

\bibitem[{Zhang et~al.(2024{\natexlab{d}})Zhang, Chen, Hu, Xu, Chen, Hao, Han, Thai, Wang, Liu et~al.}]{bench}
Xinrong Zhang, Yingfa Chen, Shengding Hu, Zihang Xu, Junhao Chen, Moo Hao, Xu~Han, Zhen Thai, Shuo Wang, Zhiyuan Liu, et~al. 2024{\natexlab{d}}.
\newblock $\infty${B}ench: Extending long context evaluation beyond 100{K} tokens.
\newblock In \emph{Proceedings of the 62nd Annual Meeting of the Association for Computational Linguistics (Volume 1: Long Papers)}, pages 15262--15277.

\bibitem[{Zhang et~al.(2023)Zhang, Li, Cui, Cai, Liu, Fu, Huang, Zhao, Zhang, Chen et~al.}]{siren}
Yue Zhang, Yafu Li, Leyang Cui, Deng Cai, Lemao Liu, Tingchen Fu, Xinting Huang, Enbo Zhao, Yu~Zhang, Yulong Chen, et~al. 2023.
\newblock Siren's song in the ai ocean: a survey on hallucination in large language models.
\newblock \emph{arXiv preprint arXiv:2309.01219}.

\bibitem[{Zheng et~al.(2024)Zheng, Zhang, Zhang, Ye, Luo, Feng, and Ma}]{llamafactory}
Yaowei Zheng, Richong Zhang, Junhao Zhang, Yanhan Ye, Zheyan Luo, Zhangchi Feng, and Yongqiang Ma. 2024.
\newblock Llamafactory: Unified efficient fine-tuning of 100+ language models.
\newblock \emph{arXiv preprint arXiv:2403.13372}.

\bibitem[{Zhu et~al.(2023)Zhu, Yang, Wang, Song, Wu, Wei, and Li}]{pose}
Dawei Zhu, Nan Yang, Liang Wang, Yifan Song, Wenhao Wu, Furu Wei, and Sujian Li. 2023.
\newblock Pose: Efficient context window extension of llms via positional skip-wise training.
\newblock \emph{arXiv preprint arXiv:2309.10400}.

\end{thebibliography}
